\documentclass{article}
\usepackage{ijcai25}

\usepackage{times}
\usepackage{multicol} 
\usepackage{soul}
\usepackage{booktabs}
\usepackage{multicol}
\usepackage{multirow}
\usepackage{url}
\usepackage{subcaption}
\usepackage[hidelinks]{hyperref}
\usepackage[utf8]{inputenc}
\usepackage{graphicx}
\usepackage{subcaption}
\usepackage{amsmath}
\usepackage{amsthm}
\usepackage{booktabs}
\usepackage{algorithm}
\usepackage{algorithmic}
\usepackage[switch]{lineno}
\usepackage{amssymb}
\usepackage{enumitem} %
\newtheorem{definition}{Definition} 
\newtheorem{assumption}{Assumption} 
\newtheorem{lemma}{Lemma}

\newtheorem{theorem}{Theorem}

\title{Learning Heterogeneous Performance-Fairness Trade-offs in Federated Learning}

\author{
Rongguang Ye \and Ming Tang\thanks{Corresponding Author.} \\
\affiliations Department of Computer Science and Engineering and the Research Institute of Trustworthy Autonomous Systems at Southern University of Science and Technology, Shenzhen, China.  \\
\emails yerg2023@mail.sustech.edu.cn, tangm3@sustech.edu.cn
}

\begin{document}

\maketitle

\begin{abstract}
Recent methods leverage a hypernet to handle the performance-fairness trade-offs in federated learning. This hypernet maps the clients' preferences between model performance and fairness to preference-specifc models on the trade-off curve, known as local Pareto front. However, existing methods typically adopt a uniform preference sampling distribution to train the hypernet across clients, neglecting the inherent heterogeneity of their local Pareto fronts. Meanwhile, from the perspective of generalization, they do not consider the gap between local and global Pareto fronts on the global dataset. To address these limitations, we propose HetPFL to effectively learn both local and global Pareto fronts. HetPFL comprises Preference Sampling Adaptation (PSA) and Preference-aware Hypernet Fusion (PHF). PSA adaptively determines the optimal preference sampling distribution for each client to accommodate heterogeneous local Pareto fronts. While PHF performs preference-aware fusion of clients' hypernets to ensure the performance of the global Pareto front. We prove that HetPFL converges linearly with respect to the number of rounds, under weaker assumptions than existing methods. Extensive experiments on four datasets show that HetPFL significantly outperforms seven baselines in terms of the quality of learned local and global Pareto fronts.
\end{abstract}

\section{Introduction}
Federated Learning (FL) \cite{mcmahan2017communication} is an emerging machine learning paradigm that designed to train neural network models using data silos while preserving data privacy. In recent years, FL has achieved remarkable success across various domains, including healthcare \cite{rieke2020future}, fintech \cite{imteaj2022leveraging}, and the Internet of Things (IoT) \cite{nguyen2021federated}. As FL continues to develop, the issue of group fairness has become an increasingly significant focus. Specifically, there are two primary types of group fairness in FL: client-based fairness \cite{li2019fair,lyu2020collaborative,wang2021federated} and group-based fairness \cite{yue2023gifair,deng2020distributionally}. Client-based fairness aims to minimize the variance in model performance across clients while preserving the overall model performance. Our work focuses on group-based fairness, which ensures that a model performs equitably across different demographic subgroups (e.g., male or female) within each local dataset \cite{kamishima2012fairness,roh2020fairbatch}.
\begin{figure}[t]
    \centering
    \setlength{\abovecaptionskip}{0.1cm}
    \includegraphics[width=\linewidth]{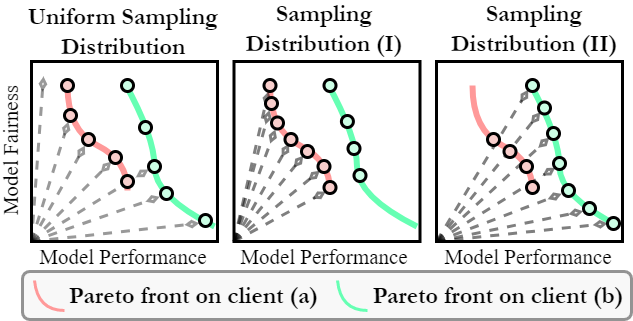}
    \caption{The impact of different sampling distributions under two clients. The dotted vectors represent preferences for model's performance and fairness. The pink and green points are loss vectors of the model after evaluation on the local dataset on client (a) and client (b), respectively. A uniform preference sampling distribution cannot achieve the best result of learning local Pareto fronts based on Lemma \ref{prop}. Instead, sampling distribution (I) is suitable for client (a), and sampling distribution (II) is suitable for client (b).}
    \label{sampling}
\end{figure}

Recent studies have focused on improving the group-based fairness of FL by proposing data sampling strategies and designing new optimization objectives. In terms of data sampling strategies, FedFB \cite{zeng2021improving} adjusts the sampling probabilities of subgroup samples during training, increasing the probability of underperforming groups to achieve group fairness. Meanwhile, FairFed \cite{ezzeldin2023fairfed} adopts a fairness-aware model aggregation scheme. Regarding the design of optimization objectives, LFT+FedAvg \cite{zeng2023federated} incorporates group fairness as a constraint during local training. FAIR-FATE \cite{salazar2023fair} introduces a linear combination of model performance and group fairness as its objective function. In fact, a trade-off exists between model performance and model fairness, meaning that improving fairness often comes at the cost of performance. With respect to efficiency, learning the entire performance-fairness trade-off curve (i.e., the Pareto front) for each client offers a more flexible scheme. Moreover, in terms of generalization, it is crucial to consider the quality of the global Pareto front on the global dataset. Recent studies \cite{linpareto,ye2024praffl} have introduced hypernets to learn the local Pareto front on the local dataset by modeling the mapping from a predefined preference distribution to preference-specific models. However, these methods still exhibit limitations in efficiency and generalization within the FL context:
\begin{itemize}
    \item \textbf{Efficiency}: 
As shown in Fig. \ref{sampling}, the heterogeneity of data in FL results in differences in the positions of local Pareto fronts across clients. Each local Pareto front has its own optimal preference sampling distribution. However, prior approaches assume a uniform preference sampling distribution across clients, which is inefficient for learning local Pareto fronts.
    \item \textbf{Generalization}: 
Achieving both optimal local and global Pareto fronts presents inherent conflicts. Prior approaches primarily focus on improving the local Pareto front while neglecting the global Pareto front.
\end{itemize}
Addressing these limitations presents several challenges. In terms of efficiency, determining the optimal preference sampling distribution for each client is non-trivial, as the position of each client's local Pareto front (ground truth) is initially unknown. Regarding generalization, aggregating clients' models to construct the optimal global model is difficult, as it requires identifying the strengths of each client's model for different preferences while maintaining data privacy.

In this paper, we propose HetPFL to efficiently learn both local and global Pareto fronts. HetPFL consists of Preference Sampling Adaptation (PSA) and Preference-aware Hypernet Fusion (PHF). PSA dynamically adjusts the preference sampling distribution by introducing data-driven HyperVolume Contribution (HVC), which quantifies each preference's contribution to the learned Pareto front. We then jointly optimize the preference sampling distribution based on HVC and the client's model as a bi-level optimization problem to enhance local Pareto front learning efficiency. To improve the global Pareto front, PHF considers a preference-aware hypernet aggregation at the server by identifying the capability of each client's hypernet for various preferences. 

The primary contributions of this work are as follows:
\begin{itemize}
\item We propose a HetPFL framework that efficiently learns the heterogeneous local Pareto fronts across clients using PSA, while simultaneously achieving a high-quality global Pareto front through PHF;
\item  We analyze the convergence rate of HetPFL within the FL system and establish an error convergence rate of order $\mathcal{O}\left(\frac{1}{t}\right)$. This result is particularly challenging to derive due to the interdependence of the different components in the FL system;
\item Extensive experiments on four datasets demonstrate that HetPFL outperforms the best-performing baseline, achieving approximately 1.75\% and 5.5\% improvements in the quality of the learned local and global Pareto fronts, respectively.
\end{itemize}

\section{Problem Formulation}
Let $\boldsymbol{x} \in \mathcal{X}$, $y \in \mathcal{Y}$ denote features and label, respectively. The features in $\boldsymbol{x} \triangleq (\boldsymbol{a}, \boldsymbol{b})$ contain sensitive features $\boldsymbol{a}$ (e.g., gender, race) and non-sensitive features $\boldsymbol{b}$. Suppose there are $K$ clients, and each client $k \in [K]$ has a local dataset, denoted by $\mathcal{D}_k$. The classifier $f_{\boldsymbol{\theta}_k}$ with learnable parameters $\boldsymbol{\theta}_k$ of client $k$ outputs a prediction $f_{\boldsymbol{\theta}_k}(\boldsymbol{x})$ from an input data $\boldsymbol{x}$.

According to \cite{zeng2021improving}, we define loss functions for model performance and fairness, respectively. Usually, model performance is characterized using cross-entropy loss
\begin{equation}\label{eq1}
\small
    \! \ell_{CE}(\boldsymbol{x},y \! \mid \! f_{\boldsymbol{\theta}_k})\!=\!-[y\log (f_{\boldsymbol{\theta}_k}(\boldsymbol{x}))\! +\!(1-y)\log (1-f_{\boldsymbol{\theta}_k}(\boldsymbol{x}))]. \!
\end{equation}
We use following loss function to quantify model fairness:
\begin{equation}\label{eq2}
     \ell_{F}(\boldsymbol{x},y  \mid  f_{\boldsymbol{\theta}_k})=\left[(\boldsymbol{a}-\bar{\boldsymbol{a}}_k)(f_{\boldsymbol{\theta}_k}(\boldsymbol{x})-\bar{f}_{\boldsymbol{\theta}_k}(\boldsymbol{x}))\right], 
\end{equation}
where $\bar{\boldsymbol{a}}_k$ and $\bar{f}_{\boldsymbol{\theta}_k}(\boldsymbol{x})$ represent the average values of $\boldsymbol{a}$ and $f_{\boldsymbol{\theta}_k}(\boldsymbol{x})$ over $\mathcal{D}_k$, respectively. $\ell_F$ measures the correlation between sensitive features and model
predictions. When the model prediction exhibits a stronger correlation with sensitive features, the value of $\ell_F$ rises, signaling a reduced model fairness. 

The trade-off between $\ell_{CE}$ and $\ell_{F}$ is quantified using a preference vector $\boldsymbol{\lambda} \in \Lambda=\{\boldsymbol{\lambda} \! \in \mathbb{R}_+^2 \! \mid \! \sum_{i=1}^2 \lambda_i = 1 \}$.
Following \cite{ye2024praffl}, we introduce a hypernet $h_{\boldsymbol{\beta}_k}:\mathbb{R}^{|\boldsymbol{\lambda}|}\rightarrow \mathbb{R}^{|\boldsymbol{\theta}_k|}$ with learnable parameters $\boldsymbol{\beta}_k$, which maps a preference vector $\boldsymbol{\lambda}$ to a preference-specific model $\boldsymbol{\theta}_k = h_{\boldsymbol{\beta}_k}(\boldsymbol{\lambda})$. We aim at optimizing $\boldsymbol{\beta}_k$ to improve model performance and fairness (i.e., reducing the losses in  Eqs. (\ref{eq1}) and (\ref{eq2})), for which we can define a \textit{weighted Tchebycheff scalar} loss \cite{miettinen1999nonlinear} for each preference vector $\boldsymbol{\lambda}$:
\begin{equation}\label{tch}
\small
\!\! \min_{\boldsymbol{\beta}_{k}} g_{\mathrm{tch}}(\boldsymbol{x},y,h_{\boldsymbol{\beta}_{k}}(\boldsymbol{\lambda})\mid \boldsymbol{\lambda})\!=\! \max_{j \in \{CE, F\}} \! \left\{\frac{\ell_j(\boldsymbol{x}, y \mid h_{\boldsymbol{\beta}_{k}}(\boldsymbol{\lambda})}{\lambda_j} \right\}. \!
\end{equation}
Eq. (\ref{tch}) satisfies the following Lemma \cite{miettinen1999nonlinear}:
\begin{lemma}[Preference Alignment]\label{prop}
    Given a preference vector $\boldsymbol{\lambda}$, a preference-specific model $h_{\boldsymbol{\beta}_{k}}(\boldsymbol{\lambda})$ is weakly Pareto optimal to the problem (\ref{tch}) if and only if $h_{\boldsymbol{\beta}_{k}}(\boldsymbol{\lambda})$ is optimal for problem (\ref{tch}). 
\end{lemma}
We give a schematic diagram of weakly Pareto optimality and Pareto optimality in Fig. 6 of the Appendix. Lemma \ref{prop} guarantees that when $h_{\boldsymbol{\beta}_{k}}(\boldsymbol{\lambda})$ is optimal for problem (\ref{tch}), the loss vector $(\ell_{CE}, \ell_{F})$ of $h_{\boldsymbol{\beta}_{k}}(\boldsymbol{\lambda})$ on dataset $\mathcal{D}_k$ aligns exactly with the direction of the preference vector and lies on the Pareto front, as shown by the points in Fig. \ref{sampling}. 
\begin{figure*}[t]
    \centering
    \setlength{\abovecaptionskip}{0.1cm}
    \includegraphics[width=0.95\linewidth]{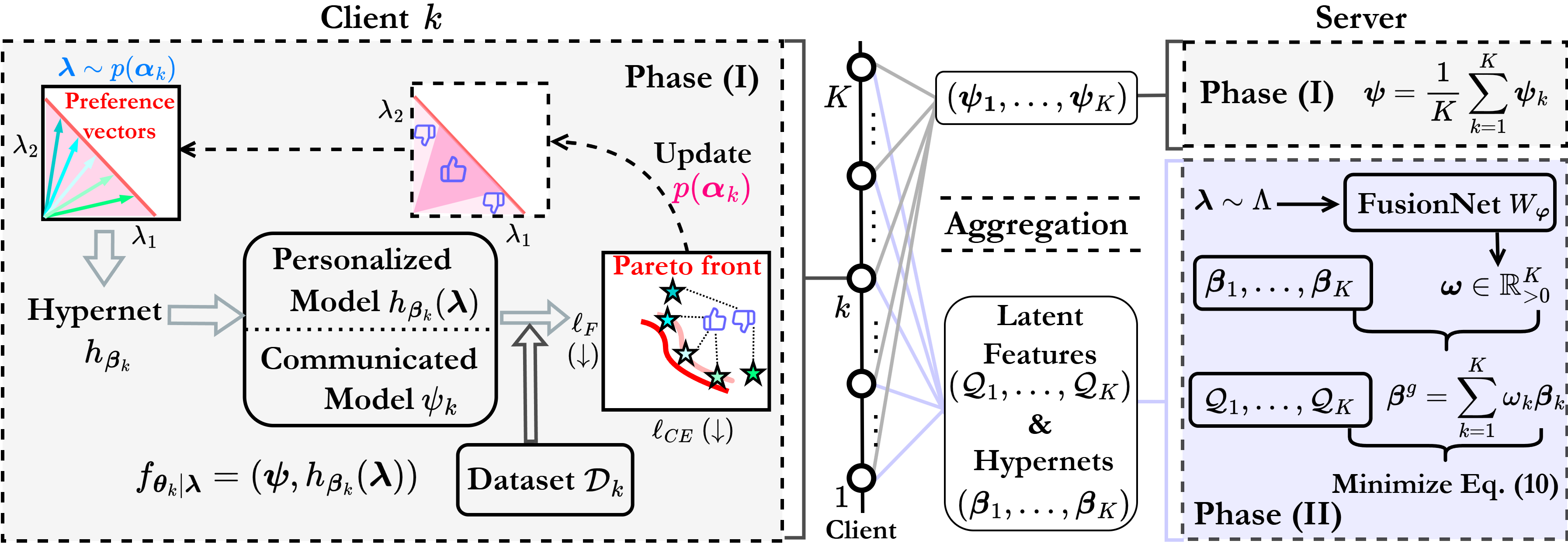}
    \caption{HetPFL framework.}
    \label{pro}
    \vspace{-0.05cm}
\end{figure*}

We consider optimizing Eq. (\ref{tch}) over the preference distribution $\Lambda_k$ of each client $k$, and define the following goal for the \textbf{\textit{local Pareto front}} learning of client $k$.
\begin{equation}\label{local2}
    \min_{\boldsymbol{\beta}_k} \mathbb{E}_{\boldsymbol{\lambda}\sim \Lambda_k}\mathbb{E}_{(\boldsymbol{x},y)\in\mathcal{D}_k}g_{\mathrm{tch}}(\boldsymbol{x},y,h_{\boldsymbol{\beta}_{k}}(\boldsymbol{\lambda})\mid \boldsymbol{\lambda}),
\end{equation}
where $\Lambda_k$ is unknown in advance and depends on the position of the Pareto front of client $k$. Preference vector $\boldsymbol{\lambda}$ is sampled from $\Lambda_k$. Once Eq. (\ref{local2}) is completed, the hypernet can receive all possible preference vectors as inputs, generating a corresponding set of preference-specific models $\{\boldsymbol{\theta}_k = h_{\boldsymbol{\beta}_k}(\boldsymbol{\lambda}) \mid \boldsymbol{\lambda} \sim \Lambda_m\}$. This model set is then evaluated on the local dataset $\mathcal{D}_k$, and the evaluation results collectively form the entire local Pareto front.

Similarly, the goal for the \textbf{\textit{global Pareto front}} is to generate an aggregated hypernet $\boldsymbol{\beta}^g = \frac{1}{K} \sum_{k=1}^K \boldsymbol{\beta}_k$, which minimizes the $g_{\mathrm{tch}}(\cdot)$ over the 
global dataset:
\begin{equation}\label{glo2}
    \min_{\boldsymbol{\beta}^g} \frac{1}{K}\sum_{k=1}^K\mathbb{E}_{\boldsymbol{\lambda}\sim \Lambda}\mathbb{E}_{(\boldsymbol{x},y)\in\mathcal{D}_k}g_{\mathrm{tch}}(\boldsymbol{x},y,h_{\boldsymbol{\beta}^g}(\boldsymbol{\lambda})\mid \boldsymbol{\lambda}),
\end{equation}
where $\Lambda$ can be an arbitrary distribution or defined as a combination of $\Lambda_1,...,\Lambda_K$.

Our goal is to optimize Eq. (\ref{local2}) for all clients while simultaneously optimizing Eq. (\ref{glo2}). The primary challenge in Eq. (\ref{local2}) arises from the distinctiveness of each client's local Pareto front. This implies the necessity of approximating preference distribution $\Lambda_k$ for each client $k$ to effectively learn the local Pareto fronts. For Eq. (\ref{glo2}), the optimization objectives for local and global Pareto fronts are inherently conflicting, preventing simultaneous optimality for both.
\section{Methodology}
This section presents our HetPFL framework. We provide an overview and introduce the two main components, PSA and PHF, of HetPFL in Sections \ref{pfl}--\ref{pha}. Sections \ref{hetalg}--\ref{theo} describe the optimization procedure of HetPFL and analyze its convergence properties.
\subsection{Overview}\label{pfl}
Fig. \ref{pro} shows our proposed HetPFL. The foundational components of HetPFL include communicated model $\boldsymbol{\psi}$, hypernet $h_{\boldsymbol{\beta}_k}$, preference sampling distribution $p(\boldsymbol{\alpha}_k)$, and FusionNet $W_{\boldsymbol{\varphi}}$. HetPFL can be divided into two phases. Phase (I) focuses on efficiently learning the local Pareto fronts for all clients, while Phase (II) aims to learn the global Pareto front.

\subsubsection{Phase (I)}
In this phase, we aim to optimize the hypernet, sampling distribution, and the communicated model. The communicated model $\boldsymbol{\psi}$ transforms the features into $d$-dimensional latent features and is periodically aggregated at the server, as in FL. The hypernet $h_{\boldsymbol{\beta}_k}$ is kept locally at the client in Phase (I). Its role is to map an arbitrary preference vector $\boldsymbol{\lambda}$ into a preference-specific model $h_{\boldsymbol{\beta}_{k}}(\boldsymbol{\lambda})$, and then $h_{\boldsymbol{\beta}_{k}}(\boldsymbol{\lambda})$ transforms the $d$-dimensional latent features to label. For each client $k$, we denote $f_{\boldsymbol{\theta}_{k}\mid \boldsymbol{\lambda}}=(\boldsymbol{\psi},h_{\boldsymbol{\beta}_{k}}(\boldsymbol{\lambda}))$. 

Previous works simply set preference sampling distribution $p(\boldsymbol{\alpha}_k)$ to be uniform across all clients. In contrast, we consider jointly optimizing the hypernet $h_{\boldsymbol{\beta}_k}$ and $p(\boldsymbol{\alpha}_k)$: 
\begin{equation}\label{local}
    \min_{\boldsymbol{\beta}_{k},\boldsymbol{\alpha}_k} \mathbb{E}_{\boldsymbol{\lambda}\sim p(\boldsymbol{\alpha}_k)}\mathbb{E}_{(\boldsymbol{x},y)\in\mathcal{D}_k}g_{\mathrm{tch}}(\boldsymbol{x}, y, f_{\boldsymbol{\theta}_{k}\mid \boldsymbol{\lambda}}),
\end{equation}
where $\boldsymbol{\alpha}_k$ are the parameters of $p(\boldsymbol{\alpha}_k)$. HetPFL improves the efficiency of learning local Pareto fronts by identifying a suitable $p(\boldsymbol{\alpha}_k)$ for each client over the preference space. Meanwhile, the communicated model optimized to improve the model performance of all clients:
\begin{equation}\label{comm}
    \min_{\boldsymbol{\psi}} \frac{1}{K}\sum_{k=1}^{K} \mathbb{E}_{(\boldsymbol{x},y)\in\mathcal{D}_k}\left[\ell_{CE}(\boldsymbol{x},y, f_{\boldsymbol{\theta}_{k}\mid \widetilde{\boldsymbol{\lambda}}}) \right],
\end{equation}
where $f_{\boldsymbol{\theta}_{k}\mid \widetilde{\boldsymbol{\lambda}}}=(\boldsymbol{\psi}, h_{\boldsymbol{\beta}_k}(\widetilde{\boldsymbol{\lambda}}))$ and $\widetilde{\boldsymbol{\lambda}}$ is a predefined preference vector. The details of update steps for Eqs. (\ref{local}) and (\ref{comm}) are provided in Section \ref{hetalg}.
\subsubsection{Phase (II)}
Unlike previous works that overlook the global Pareto front, Phase (II) focuses on addressing this gap. After the final round, clients first transmit their hypernets and latent features, $\mathcal{Q}_k = {\boldsymbol{\psi}(\boldsymbol{x}) \mid (\boldsymbol{x}, y) \in \mathcal{D}k}$, to the server. The server then uses FusionNet, $W{\boldsymbol{\varphi}}$, with parameters $\boldsymbol{\varphi}$ to learn effective aggregation strategies for these hypernets, tailoring the aggregation process to different preference vectors. Notably, transmitting only the latent features of the dataset is a common practice to mitigate privacy concerns \cite{thapa2022splitfed}.

Given this framework, two key questions remain to be addressed. First, how can $p(\boldsymbol{\alpha}_k)$ be determined in Phase (I) to enable efficient learning of local Pareto fronts? Second, how to optimize FusionNet in Phase (II)? These questions will be explored in the following two subsections.
\subsection{Preference Sampling Adaptation}\label{psa}
In Phase (I), we propose Preference Sampling Adaptation (PSA) to determine $p(\boldsymbol{\alpha}_k)$. Determining $p(\boldsymbol{\alpha}_k)$ involves optimizing the quality of the sampled preference vectors during training. This process is non-trivial and unfolds in two steps: first, evaluating the quality of the sampled preference vectors without a true Pareto front (ground truth); and second, integrating the optimization of $p(\boldsymbol{\alpha}_k)$ into the hypernet's training.
 
\begin{figure}[t]
    \centering
    \setlength{\abovecaptionskip}{0.06cm}
    \includegraphics[width=0.95\linewidth]{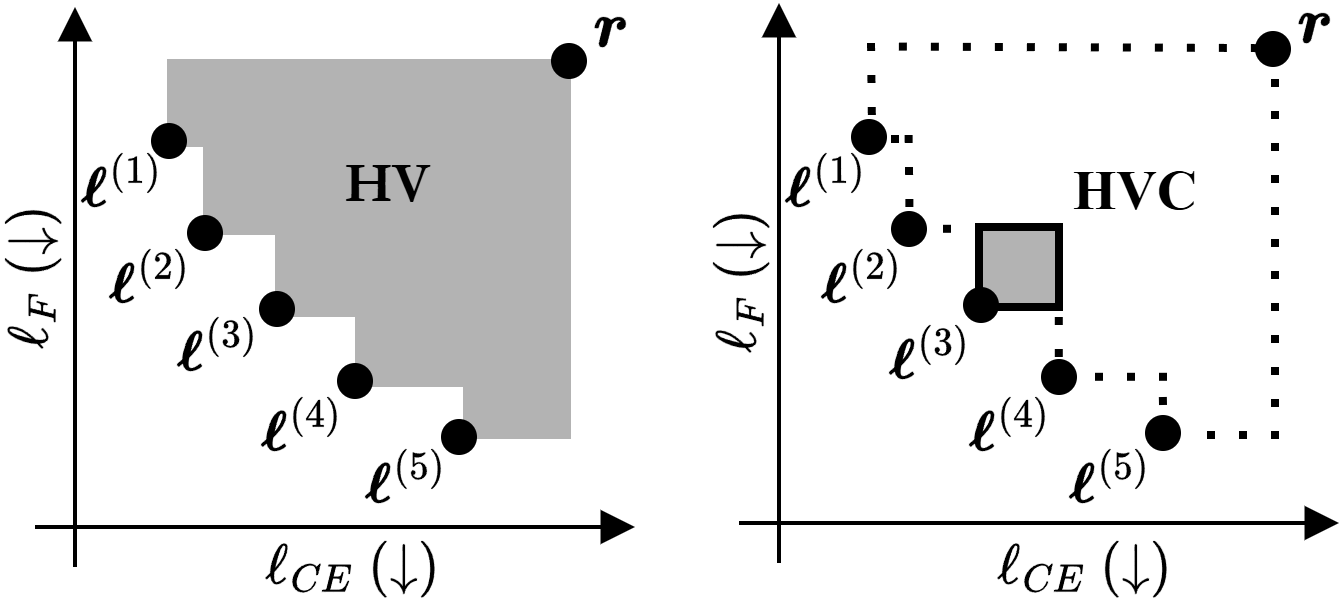}
    \caption{An illustration of the HV and HVC of the loss vectors $\boldsymbol{\ell}^{(i)}$ produced by evaluating a model set $\{f_{\boldsymbol{\theta}_k \mid {\boldsymbol{\lambda}^{i}}}\}_{i=1}^{5}$ corresponding to five input preference vectors.}
    \label{hv-hvc}
\end{figure}

For the first step, we introduce a data-driven Hypervolume Contribution (HVC) indicator to assess the quality of sampled preference vectors. Despite the absence of a true local Pareto front, it allows for quantifying each preference vector’s contribution based on the training losses.
\begin{definition}[HVC]
    Given a reference point $\boldsymbol{r}$. Let $\mathcal{S}(\boldsymbol{\lambda}, \boldsymbol{r}) = \left\{ \boldsymbol{q} \in \mathbb{R}^2 \mid \boldsymbol{\ell}(\boldsymbol{x}, y \mid f_{\boldsymbol{\theta}_{k\mid \boldsymbol{\lambda}}}) \leq \boldsymbol{q} \ \text{and} \ \boldsymbol{q} \leq \boldsymbol{r} \right\}$, and the hypervolume of a set of $N$ preference vectors \( \Lambda_{\boldsymbol{\alpha}_k}=\{\boldsymbol{\lambda}^1,...,\boldsymbol{\lambda}^N \mid \boldsymbol{\lambda}^i \sim p(\boldsymbol{\alpha}_k)\} \) is
    \[
    \small
    \mathcal{H}_{\boldsymbol{r}}\left(\Lambda_{\boldsymbol{\alpha}_k} \right) = \mathcal{L}\left(\bigcup_{\boldsymbol{\lambda} \in \Lambda_{\boldsymbol{\alpha}_k}} \mathcal{S}(\boldsymbol{\lambda}, \boldsymbol{r})\right),
    \]
    where $\mathcal{L}(\cdot)$ denotes the Lebesgue measure and $\boldsymbol{q}$ represents any point in the gray area in the left figure of Fig. \ref{hv-hvc}. The HVC of $f_{\boldsymbol{\theta} \mid {\boldsymbol{\lambda}^{i}}}$ for the set \( \Lambda_{\boldsymbol{\alpha}_k} \) is the difference between $\mathcal{H}_{\boldsymbol{r}}(\Lambda_{\boldsymbol{\alpha}_k})$ and $\mathcal{H}_{\boldsymbol{r}}(\Lambda_{\boldsymbol{\alpha}_k} \setminus {\boldsymbol{\lambda}^i})$, as follows:
\[\mathcal{HC}_{\boldsymbol{r}}({\boldsymbol{\lambda}^i}\mid \Lambda_{\boldsymbol{\alpha}_k})=\mathcal{H}_{\boldsymbol{r}}(\Lambda_{\boldsymbol{\alpha}_k}) - \mathcal{H}_{\boldsymbol{r}}(\Lambda_{\boldsymbol{\alpha}_k} \setminus {\boldsymbol{\lambda}^i}).\]    
\end{definition}

Fig. \ref{hv-hvc} shows that the HVC of $f_{\boldsymbol{\theta}_k \mid {\boldsymbol{\lambda}^{i}}}$ is the difference between the HV of the full set of five models and that of the set excluding $f_{\boldsymbol{\theta}_k\mid\boldsymbol{\lambda}^{i}}$. The larger $\mathcal{HC}_{\boldsymbol{r}}(\boldsymbol{\lambda}^i\mid \Lambda_{\boldsymbol{\alpha}_k})$, the greater the contribution of $\boldsymbol{\lambda}^i$, indicating a higher quality of $\boldsymbol{\lambda}^i$. 

We then move on to the second step. Based on HVC, we propose the following bi-level optimization objective to alternately optimize the hypernet and the preference sampling distribution of client $k$: 
\begin{small}
\begin{align}
      & \quad \quad \min_{\boldsymbol{\alpha}_k} \mathbb{E}_{\boldsymbol{\lambda} \sim p(\boldsymbol{\alpha}_k)} \mathbb{E}_{(\boldsymbol{x},y)\in \mathcal{D}_k}  [- \mathcal{HC}_{\boldsymbol{r}} \left(\boldsymbol{\lambda}\mid \Lambda_{\boldsymbol{\alpha}_k}\right)], \label{hvc_1} \\
      &\mathrm{s.t.} \  \boldsymbol{\beta}_k  =  \arg \min_{\boldsymbol{\beta}_k} \mathbb{E}_{\boldsymbol{\lambda}\sim p(\boldsymbol{\alpha}_k)} \mathbb{E}_{(\boldsymbol{x},y)\in \mathcal{D}_k} [g_{\mathrm{tch}}(\boldsymbol{x},y,f_{\boldsymbol{\theta}_{k}\mid \boldsymbol{\lambda}})], \label{hvc_2}
\end{align}
\end{small}
\!\!where $f_{\boldsymbol{\theta}_{k}\mid \boldsymbol{\lambda}}=\left(\boldsymbol{\psi}_k, h_{\boldsymbol{\beta}_k}(\boldsymbol{\lambda})\right)$. The bi-level optimization first optimizes the hypernet as in Eq. (\ref{hvc_2}). Then, the sampling distribution is further refined based on the solution of Eq. (\ref{hvc_2}). As shown in Fig. \ref{pro}, it makes the sampling distribution at the next iteration more beneficial to local Pareto front learning.
\subsection{Preference-aware Hypernet Fusion}\label{pha}
In Phase (II), we propose a preference-aware hypernet fusion (PHF) method. The intuition behind PHF is that each client's hypernet excels at specific preference vectors. Thus, for any given preference vector, PHF learns the preference-aware aggregation weight so that hypernets specializing in that vector are given higher weight, thereby improving the quality of the global Pareto front.  As shown in Fig. \ref{pro}, we introduce a FusionNet $W_{\boldsymbol{\varphi}}:\mathbb{R}^{|\boldsymbol{\lambda}|} \rightarrow \mathbb{R}^{K}_{>0}$ with parameters $\boldsymbol{\varphi}$ that learns a mapping from any preference vector $\boldsymbol{\lambda}$ to a fusion weight $\boldsymbol{\omega}=W_{\boldsymbol{\varphi}}(\boldsymbol{\lambda})\in \mathbb{R}^K$. Based on the fusion weight, the hypernets from all clients are linearly combined to form a global hypernet ${\boldsymbol{\beta}^g}=W_{\boldsymbol{\varphi}}(\boldsymbol{\lambda}) \cdot [{\boldsymbol{\beta}_1},..., {\boldsymbol{\beta}_K}]$, where operation $\cdot$ is the vector inner product. This process is optimized through 
\begin{equation}\label{glo}
    \min_{\boldsymbol{\varphi}} \mathbb{E}_{\boldsymbol{\lambda}\sim \Lambda} \left[\frac{1}{K} \sum_{k=1}^K g_{\mathrm{tch}}(\mathcal{Q}_k, \mathcal{Y}_k, f_{\boldsymbol{\theta}^g_{k} \mid \boldsymbol{\lambda}})\right],
\end{equation}
where $f_{\boldsymbol{\theta}^g_{k}\mid \boldsymbol{\lambda}}=(\boldsymbol{\psi}, h_{\boldsymbol{\beta}^g}(\boldsymbol{\lambda}))$, and $\mathcal{Y}_k$ represents labels on dataset $\mathcal{D}_k$. Recalling Lemma \ref{prop}, the global hypernet $f_{\boldsymbol{\theta}^g_{k}\mid \boldsymbol{\lambda}}$ aggregated by FusionNet is optimized towards the direction where the Pareto front intersects with $\boldsymbol{\lambda}$. Once Eq. (\ref{glo}) is solved, the mapping from preference vectors to fusion weights during inference is highly efficient.

\subsection{Algorithm: HetPFL}\label{hetalg}
In this subsection, we present HetPFL algorithm to optimizing four components including communicated model $\boldsymbol{\psi}$, hypernet $h_{\boldsymbol{\beta}_k}$, preference sampling distribution $p(\boldsymbol{\alpha}_k)$, and FusionNet $W_{\boldsymbol{\varphi}}$. At the beginning of Phase (I), each client downloads the global communicated model $\boldsymbol{\theta}_k^0$ from the server. 

In round $t$, the communicated model on client $k$ is updated through $\tau_c$ steps of gradient descent with a learning rate of $\eta_t$:
\begin{equation}\label{opcomm}
     \boldsymbol{\psi}^t \leftarrow \boldsymbol{\psi}^t- \eta_t \mathbb{E}_{(\boldsymbol{x},y)\in \mathcal{D}_k} \nabla_{\psi} \left[\ell_{CE}(\boldsymbol{x},y \mid f_{\boldsymbol{\theta}_k\mid\widetilde{\boldsymbol{\lambda}}})\right].
\end{equation}
To balance model performance and fairness, we set $\widetilde{\boldsymbol{\lambda}}$ to $(\frac{1}{2}, \frac{1}{2})$ in Eq. (\ref{opcomm}). Then, we proceed to optimize the hypernet and the preference sampling distribution (in Eqs. (\ref{hvc_2}) and (\ref{hvc_1})). Note that lower-level problem (Eq. (\ref{hvc_2})) is a stochastic optimization problem, which is challenging to solve directly due to the expectation over preference distribution involving infinite possible values. To address this, we approximate the expectation term using Monte Carlo sampling and then solve the Eq. (\ref{hvc_2}) with $\tau_p$ steps of gradient descent
\begin{equation}\label{eq10}
\small
      \!\!\!  \boldsymbol{\beta}_{k}^t \leftarrow \boldsymbol{\beta}_k^t 
         - \frac{\eta_t}{N}\mathbb{E}_{(\boldsymbol{x},y)\in \mathcal{D}_k}  
        \sum_{v=1}^{N} \nabla_{\boldsymbol{\beta}_k} g_{\mathrm{tch}}(\boldsymbol{x},y ,f_{\boldsymbol{\theta}_k\mid \boldsymbol{\lambda}^v}\! \mid  \!{\boldsymbol{\lambda}^v}),\!
\end{equation}
where $\eta_t$ denotes the learning rate, $\boldsymbol{\lambda}^v$ is a sampled preference vector and $N$ is the number of sampled preference vectors.

Solving the upper-level problem in Eq. (\ref{hvc_1}) relies on computing the HVC gradient, given by $g_{\boldsymbol{\alpha}_k}=\nabla_{\boldsymbol{\alpha}_k} \mathbb{E}_{\boldsymbol{\lambda}\in \Lambda_{\boldsymbol{\alpha}_k}} \mathbb{E}_{(\boldsymbol{x},y)\in \mathcal{D}_k}
        \left[-\mathcal{HC}_{\boldsymbol{r}}(\boldsymbol{\lambda}\mid \Lambda_{\boldsymbol{\alpha}_k})\right]$. However, since this gradient is sometimes non-differentiable, we employ Natural Evolution Strategies (NES) \cite{salimans2017evolution}, which yield gradient estimation $\hat{g}_{\boldsymbol{\alpha}_k}$ for $g_{\boldsymbol{\alpha}_k}$:
\begin{equation}\label{eq13}
    \begin{aligned}
       \! \hat{g}_{\boldsymbol{\alpha}_k}
        \approx \mathbb{E}_{\boldsymbol{\lambda}\in \Lambda_{\boldsymbol{\alpha}_k}} \left[-\mathcal{HC}_{\boldsymbol{r}}(\boldsymbol{\lambda}\!\mid\! \Lambda_{\boldsymbol{\alpha}_k}) \nabla_{\boldsymbol{\alpha_k}} \! \log p(\boldsymbol{\lambda} \! \mid \! \boldsymbol{\alpha}_k)\right],
    \end{aligned}
\end{equation}
where $\Lambda_{\boldsymbol{\alpha}_k}$ is the set of $N$ preference vectors collected in Eq. (\ref{eq10}). This gradient computation method only requires the preference sampling distribution $p(\boldsymbol{\alpha}_k)$ to be differentiable, without the need for HVC function to be differentiable. Based on Eq. (\ref{eq13}), to optimize Eq. (\ref{hvc_1}), $\boldsymbol{\alpha}_k$ is updated using the gradient $\hat{g}_{\boldsymbol{\alpha}}$ by performing $\tau_p$ steps of gradient descent:
\begin{equation}\label{opalpha}
    \boldsymbol{\alpha}_k^t \leftarrow \boldsymbol{\alpha}^t_k - \kappa_t \hat{g}_{\boldsymbol{\alpha}_k},
\end{equation}
where $\kappa_t$ is the learning rate. After round $t$ is completed, all clients transmit their communicated models $\boldsymbol{\psi}_k, k\in[K]$, to the server. The server updates the communicated model by performing an averaging aggregation
\begin{equation}
    \boldsymbol{\psi}^{t+1}_k = \frac{1}{K}\sum_{k=1}^K \boldsymbol{\psi}^{t}_k.
\end{equation}
Subsequently, each client initializes the communicated model as the aggregated model for round $t+1$. 

Upon completing a total of $T$ rounds, we proceed to Phase (II), where the optimization of $\boldsymbol{\varphi}^t$ is updated by
\begin{equation}\label{opfusion}
   \!\! \boldsymbol{\varphi}^t \leftarrow \boldsymbol{\varphi}^t - \eta_t \frac{1}{N} \frac{1}{K} \sum_{v=1}^{N} \sum_{k=1}^K \nabla_{\boldsymbol{\varphi}} g_{\mathrm{tch}}(\mathcal{Q}_k,\mathcal{Y}, f_{\boldsymbol{\theta}_k^g\mid \boldsymbol{\lambda}} \mid \boldsymbol{\lambda}), \!
\end{equation}
where $\eta_t$ is the learning rate. We provide the pseudocode of the HetPFL algorithm in the Appendix.
\subsection{Theoretical Analysis}\label{theo}
In this subsection, we theoretically analyze the convergence of HetPFL algorithm. Our proof process is structured in two main steps. Firstly, we establish an upper bound of the communicated model at any given round $t$. Next, we provide the upper bound of the hypernet at any given round $t$.

To simplify the notation, we represent the expressions of Eqs. (\ref{hvc_1}) and (\ref{hvc_2}) as $g_{\mathrm{hvc}}(\boldsymbol{\alpha}_k,\boldsymbol{\beta}_k)=\mathbb{E}_{\boldsymbol{\lambda} \sim p(\boldsymbol{\alpha}_k)} \mathbb{E}_{(\boldsymbol{x},y)\in \mathcal{D}_k}  [- \mathrm{HVC}_{\boldsymbol{r}}(\boldsymbol{\lambda}\! \mid  \Lambda_{\boldsymbol{\alpha}_k})]$ and $g_{\mathrm{tch}}(\boldsymbol{\alpha}_k,\boldsymbol{\beta}_k)=\mathbb{E}_{\boldsymbol{\lambda}\sim p(\boldsymbol{\alpha}_k)} \mathbb{E}_{(\boldsymbol{x},y)\in \mathcal{D}_k} [g_{\mathrm{tch}}(\boldsymbol{x},y,f_{\boldsymbol{\theta}_{k \! \mid\! \boldsymbol{\lambda}}}\mid \boldsymbol{\lambda})]$, respectively. 

Based on \cite{hong2023two}, we make the following assumptions.

\begin{assumption}
    $\nabla_{\boldsymbol{\beta}}g_{\mathrm{tch}}(\boldsymbol{\alpha}_k,\boldsymbol{\beta}_k)$, 
    $\nabla^2_{\boldsymbol{\alpha}\boldsymbol{\beta}}g_{\mathrm{tch}}(\boldsymbol{\alpha}_k,\boldsymbol{\beta}_k)$,
    $\nabla^2_{\boldsymbol{\beta}\boldsymbol{\beta}}g_{\mathrm{tch}}(\boldsymbol{\alpha}_k,\boldsymbol{\beta}_k)$, 
    $\nabla_{\boldsymbol{\alpha}}g_{\mathrm{hvc}}(\boldsymbol{\alpha}_k,\boldsymbol{\beta}_k)$, and 
    $\nabla_{\boldsymbol{\beta}}g_{\mathrm{hvc}}(\boldsymbol{\alpha}_k, \boldsymbol{\beta}_k)$ 
    are Lipschitz continuous in $\boldsymbol{\beta}_k$ with respective Lipschitz constants $L_{t1}, L_{t2}$, $L_{t3}$, $L_{h1}$ and $L_{h2}$.
\end{assumption}

\begin{assumption}
$\nabla^2_{\boldsymbol{\alpha}\boldsymbol{\beta}}g_{\mathrm{tch}}$$(\boldsymbol{\alpha}_k,\boldsymbol{\beta}_k)$,$\nabla^2_{\boldsymbol{\beta}\boldsymbol{\beta}}g_{\mathrm{tch}}(\boldsymbol{\alpha}_k,\boldsymbol{\beta}_k)$, $\nabla_{\boldsymbol{\beta}}g_{\mathrm{hvc}}(\boldsymbol{\alpha}_k,\boldsymbol{\beta}_k)$ is Lipschitz continuous in $\boldsymbol{\alpha}_k$ with respective Lipschitz constants $L_{t4}$, $L_{t5}$ and $L_{h3}$.
\end{assumption}

\begin{assumption}
$g_{\mathrm{tch}}(\boldsymbol{\alpha}_k, \boldsymbol{\beta}_k)$ is $\mu_1$-strongly convex in $\boldsymbol{\beta}_k$, and ${g}_{\mathrm{tch}}(\boldsymbol{\beta}_k, \boldsymbol{\alpha}^*_k)$ is $\mu_2$-strongly convex in $\boldsymbol{\beta}_k$, where $\boldsymbol{\alpha}_k^*$ is optimal sampling distribution for client $k$.
\end{assumption}

\begin{assumption}
The expectation of stochastic gradients is always bounded. That is, $||\nabla^2_{\boldsymbol{\alpha}\boldsymbol{\beta}}g_{tch}(\boldsymbol{\alpha}_k, \boldsymbol{\beta}_k)||\leq G_1$, $||\nabla_{\boldsymbol{\alpha}}g_{hvc}(\boldsymbol{\alpha}_k, \boldsymbol{\beta}_k)||\leq G_2$, and $||\nabla_{\boldsymbol{\beta}}g_{tch}(\boldsymbol{\alpha}_k, \boldsymbol{\beta}_k)||\leq G_3$.
\end{assumption}

Let $\Delta_{\boldsymbol{\beta}_k}^t \triangleq \mathbb{E}\left[||\boldsymbol{\beta}_k^t-\boldsymbol{\beta}_{k\mid {\boldsymbol{\alpha}_k^{t-1}}}^*||^2\right]$ denote the error between the hypernet at round $t$ and the optimal hypernet $\boldsymbol{\beta}_{k\mid {\boldsymbol{\alpha}_k^{t-1}}}^*$ given the sampling distribution $p(\boldsymbol{\alpha}_k^{t-1})$ in round $t-1$. Let $\Delta_{\boldsymbol{\psi}_k}^t \triangleq \mathbb{E}\left[||\boldsymbol{\psi}_k^t-\boldsymbol{\psi}_k^*||^2\right]$ denote the error between the communicated model at round $t$ and the optimal communicated model $\boldsymbol{\psi}_k^*$. The communicated model has following upper bound.
\begin{lemma}[Convergence of the Communicated Model \cite{collins2021exploiting}]\label{commconv}
If the communicated model $\boldsymbol{\psi}$ is optimized by FedAvg \cite{mcmahan2017communication} and given a constant $\zeta > 0$, then $\boldsymbol{\psi}$ converges to the optimal communicated model $\boldsymbol{\psi}^*$ at a linear rate:
\begin{equation}
\Delta_{\boldsymbol{\psi}_k}^t \leq (1-\eta \zeta)^{t/2} \ \Delta_{\boldsymbol{\psi}_k}^0,
\end{equation}
with a probability at least $1 - t e^{-100 \min(|\boldsymbol{x}|^2\log(|K|), d)}$. 
\end{lemma}
Lemma \ref{commconv} shows that the error convergence rate is $\mathcal{O}(\frac{1}{t})$. Under weaker assumptions compared to \cite{ye2024praffl} (i.e., without requiring the initial convergence error of the hypernet to be a constant multiple of the communicated model), we establish the following upper bound for hypernet. The detailed proof can be found in the Appendix.

\begin{theorem}[Convergence of the Hypernet]\label{them}
    Under Assumptions 1-4 and Lemma \ref{commconv}, the upper bound of hypernet is
        \begin{small}
        \begin{equation}
        \begin{aligned}
            \Delta_{\boldsymbol{\beta}_k}^{t+1} \leq 
            & \ (\frac{3}{4})^{\tau_pt}\Delta_{\boldsymbol{\beta}_k}^0 + z_1(1-\eta_t\zeta)^{t/4}\sqrt{\Delta^0_{\boldsymbol{\psi}_k}}
            \\ &+
            z_2(1-\eta_t\zeta)^{t/2}\Delta^0_{\boldsymbol{\psi}_k}+ 
    \frac{\sigma_1^2\mu_1+c_1^2L^2_{q1}+G_3^2\mu_1}{\mu^3_1}
            \\ &+
            2\eta_tL_{t1}(1-\eta_t\zeta)^{t/4}\sqrt{\Delta^0_{\boldsymbol{\psi}_k}\Delta^0_{\boldsymbol{\beta}_k}},
            \end{aligned}
        \end{equation}
    \end{small}
\end{theorem}
\noindent{where} $z_1, z_2$ are constants, and $\Delta_{\boldsymbol{\alpha}_0}^t = \mathbb{E}\left[||\boldsymbol{\alpha}_k^0-\boldsymbol{\alpha}_k^{*}||^2\right]$. Theorem \ref{them} guarantees an optimization error of order $\mathcal{O}((\frac{3}{4})^{\tau_p t}+(1-\eta_t\zeta)^{t/4}+(1-\eta_t\zeta)^{t/2}+\frac{\sigma_1^2\mu_1+c_1^2L^2_{q1}+G_3^2\mu_1}{\mu^3_1})$. When $t \rightarrow +\infty$, $\Delta_{\boldsymbol{\beta}_k}^{t+1}$ converges to $ 
    \frac{\sigma_1^2\mu_1+c_1^2L^2_{q1}+G_3^2\mu_1}{\mu^3_1}$. Due to $(1-\eta_t\zeta)^{t/4}$ being the dominant term in the error convergence rate, the overall error convergence rate is $\mathcal{O}(\frac{1}{t})$.
\section{Experiments}
\begin{table*}[t]
    \centering
    \setlength{\abovecaptionskip}{0.05cm}
    \caption{Averaged performance comparison of different methods across four datasets over three runs. The best results are highlighted in bold, while the second-best results are underlined.}\label{main}
    \setlength{\tabcolsep}{5pt}
    \begin{tabular}{lcccccccc}
        \toprule
        \multirow{2.5}{*}{Method} & \multicolumn{2}{c}{SYNTHETIC} & \multicolumn{2}{c}{COMPAS} & \multicolumn{2}{c}{BANK} & \multicolumn{2}{c}{ADULT} \\
        \cmidrule(lr){2-3} \cmidrule(lr){4-5} \cmidrule(lr){6-7} \cmidrule(lr){8-9}
        & Local HV & Global HV & Local HV & Global HV & Local HV & Global HV & Local HV & Global HV \\
        \midrule
        LFT+Ensemble & 0.425 & 0.479 & 0.514 & 0.555 & 0.890 & 0.881 & 0.760 & 0.764 \\
        LFT+Fedavg & 0.700 & 0.468 & 0.505 & 0.514 & 0.891 & 0.138 & 0.765 & 0.501 \\
        Agnosticfair & 0.492 & 0.537 & 0.499 & 0.550 & 0.887 & 0.880 & \underline{0.780} & \underline{0.783} \\
        FairFed & 0.339 & 0.367 & 0.418 & 0.434 & 0.889 & 0.878 & 0.267 & 0.270 \\
        FedFB & 0.567 & 0.608 & 0.505 & 0.517 & 0.893 & 0.883 & 0.759 & 0.763 \\
        EquiFL & 0.642 & 0.604 & 0.564 & 0.526 & 0.892 & 0.882 & 0.761 & 0.764 \\ 
        PraFFL & \underline{0.800} & \underline{0.716} & \underline{0.599} & \underline{0.613} & \underline{0.901} & \underline{0.895} & 0.766 & 0.750 \\
        HetPFL (Ours) & \textbf{0.830} & \textbf{0.827} & \textbf{0.623} & \textbf{0.626} & \textbf{0.904} & \textbf{0.898} & \textbf{0.783} & \textbf{0.846} \\
        \bottomrule
    \end{tabular}
\end{table*}
\begin{figure*}[t]
    \centering
    \setlength{\abovecaptionskip}{0.06cm}
    \includegraphics[width=\linewidth]{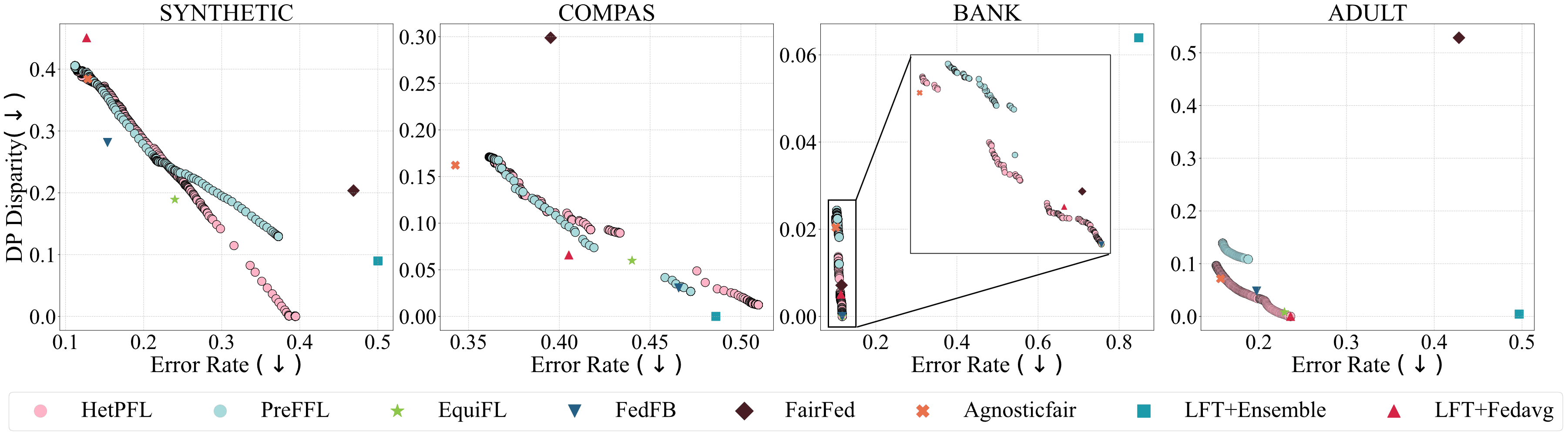}
    \caption{Comparison of global Pareto front obtained by our HetPFL algorithm and baselines on four datasets. A Pareto front closer to the bottom-left corner indicates better performance.}
    \label{pf}
\end{figure*}

\subsection{Experimental Settings}
\noindent{\textbf{Datasets.}}
Four widely-used datasets are employed to evaluate the performance of HetPFL, including a SYNTHETIC \cite{zeng2021improving}, COMPAS \cite{barenstein2019propublica}, BANK \cite{moro2014data}, and ADULT \cite{dua2017uci}.

\noindent{\textbf{Baselines.}}
We compare HetPFL with seven state-of-the-art methods, including two for addressing local fairness (LFT+Ensemble and LFT+Fedavg \cite{zeng2023federated}), three for global fairness (Agnosticfair \cite{du2021fairness}, FairFed \cite{ezzeldin2023fairfed} and FedFB \cite{zeng2021improving}), one for both local and global fairness (\cite{makhija2024achieving}), and one for learning performance-fairness local Pareto fronts (PraFFL \cite{ye2024praffl}). PraFFL is the most closely related work to ours in learning Pareto fronts.

\noindent{\textbf{Metrics}.} Based on \cite{ezzeldin2023fairfed}, we use the model's error rate to quantify its performance and the DP disparity \cite{feldman2015certifying} to measure its fairness, where a smaller DP disparity indicates a fairer model. Each point on the Pareto front represents a specific trade-off between performance and fairness in the two-dimensional objective space. Additionally, the hypervolume (HV) \cite{zitzler1999multiobjective} is employed to evaluate the quality of the learned Pareto front. We mainly report local HV on local datasets and global HV on global datasets in our experiments.

\noindent{\textbf{Hyperparameters}.} Since PraFFL and HetPFL have the ability to generate any number of models during inference, we set them to generate 1,000 preference-specific models each for evaluation. See the Appendix for other detailed settings. Our implementation is available at \href{https://github.com/rG223/HetPFL}{\texttt{{https://github.com/rG223/HetPFL}}}.

\subsection{Experimental Results}
We first present the main results and convergence analysis, followed by performance analysis across different scenarios, such as different levels of data heterogeneity and large-scale clients. Finally, an ablation study analyzing the two components of HetPFL is provided. 
\noindent{\textbf{Main Results.}}
We draw two key conclusions based on Table \ref{main}: (I) Methods capable of learning the Pareto front, such as PraFFL and HetPFL, outperform those that do not in most cases, in terms of both local HV and global HV. Fig. \ref{pf} shows that PraFFL and HetPFL are capable of generating more models that caters to large-scale preferences; (II) PraFFL focuses primarily on learning local Pareto fronts, it neglects the heterogeneity of Pareto fronts across clients and fails to ensure the quality of the global Pareto front, leading suboptimal on most datasets. In comparison, HetPFL outperforms PraFFL in terms of local HV and global HV across four datasets, achieving varying degrees of improvement. Fig. \ref{pf} indicates that HetPFL outperforms PraFFL on SYNTHETIC, BANK, and ADULT datasets. However, on the COMPAS, the Pareto front splits into two segments with a 3\% disconnection in terms of DP disparity and error rate. HetPFL’s unimodal sampling distribution prioritizes the tail regions' benefits while overlooking the middle sections compared to PraFFL.

\begin{figure}[t]
    \centering
    \setlength{\abovecaptionskip}{0.06cm}
    \includegraphics[width=0.495\linewidth,page=1]{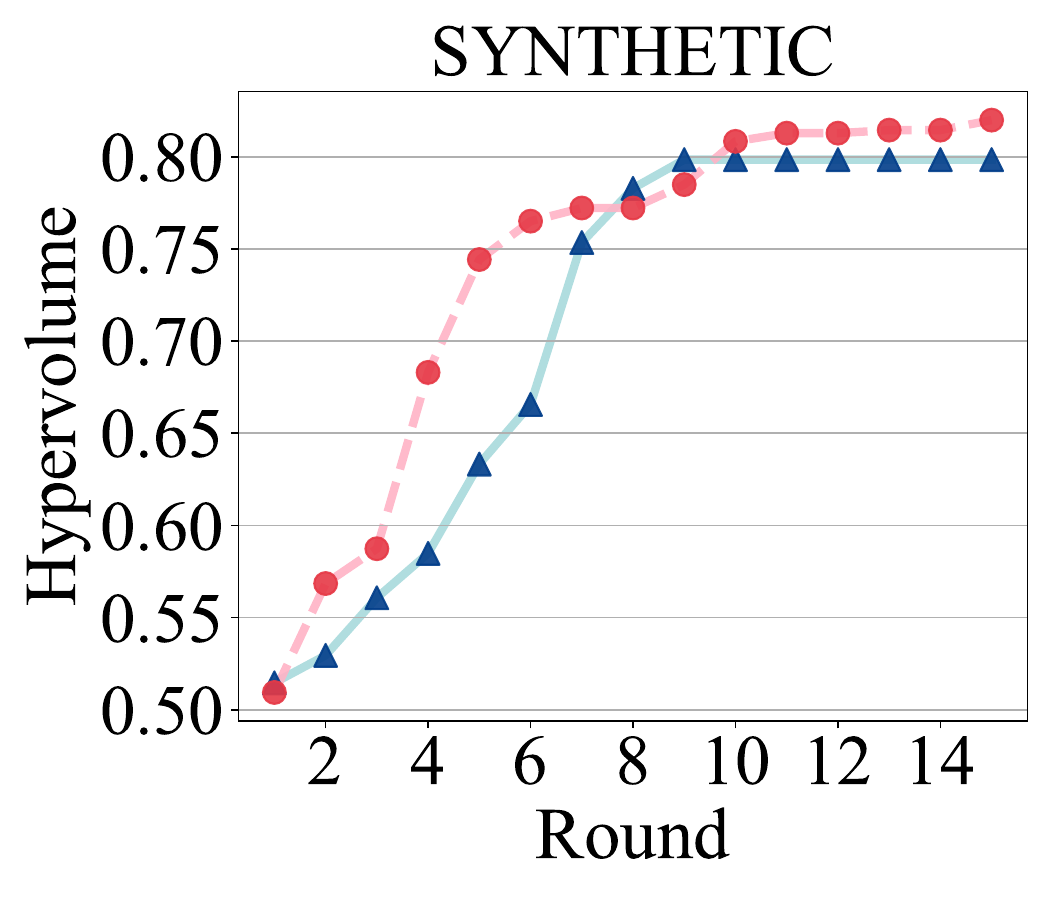} 
    \includegraphics[width=0.495\linewidth,page=1]{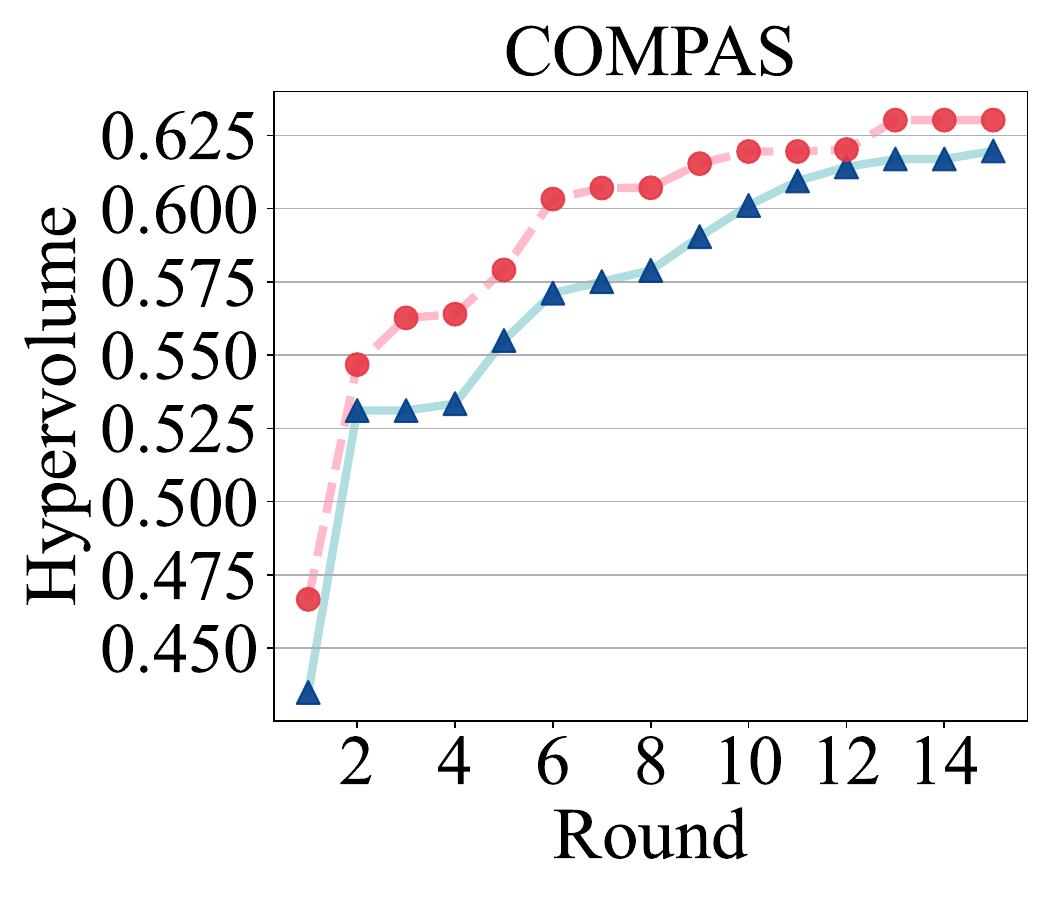} 
    \\
    \includegraphics[width=0.495\linewidth,page=1]{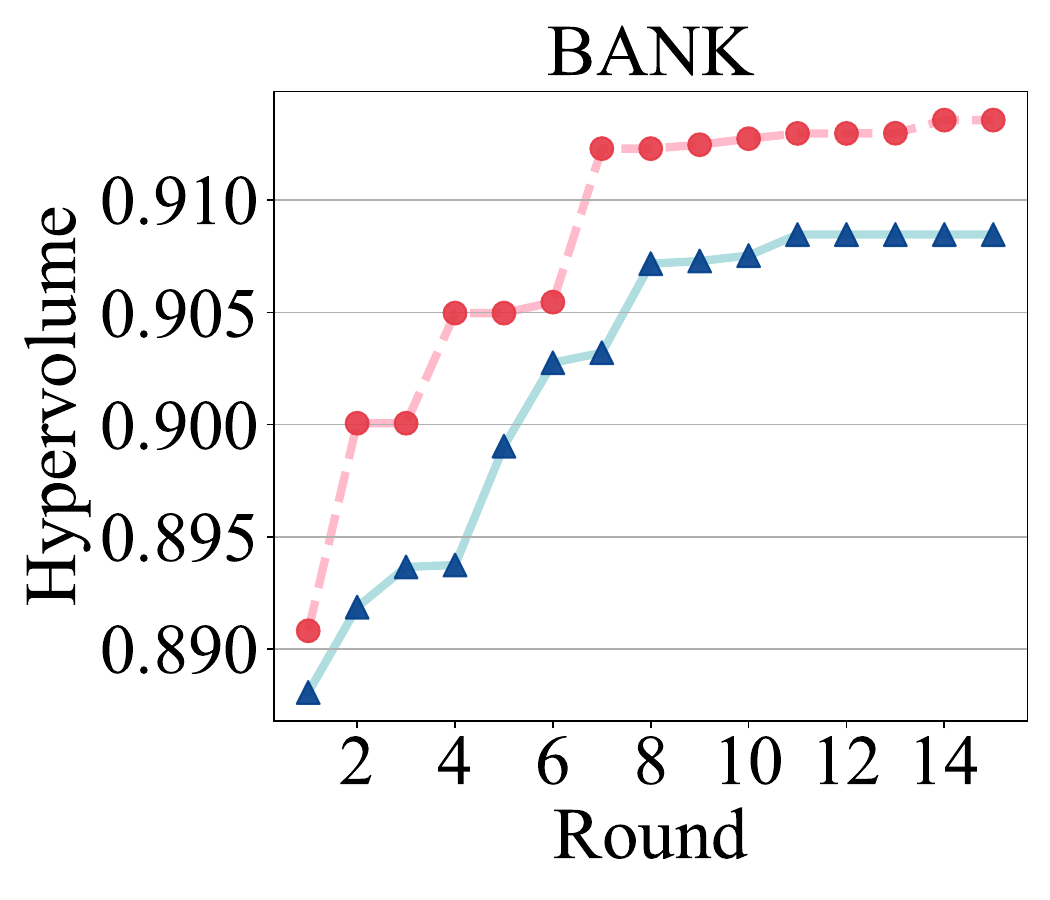} 
    \includegraphics[width=0.495\linewidth,page=1]{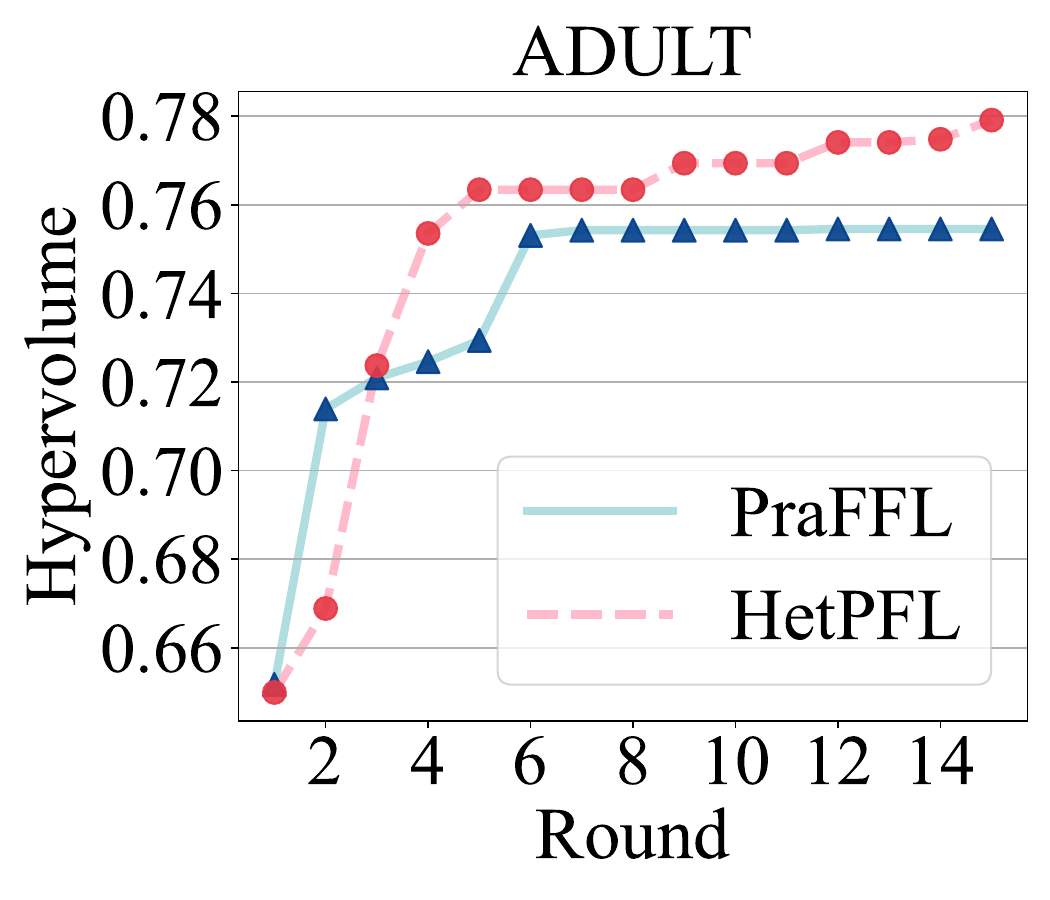} 
        \caption{Convergence of HetPFL compared with PraFFL.}
        \vspace{-0.2cm}
    \label{conv2}
\end{figure}

\noindent{\textbf{Convergence Results.}}
Fig. \ref{conv2} shows the convergence comparison of PraFFL and HetPFL on the client local validation set in each round. HetPFL shows consistently faster convergence compared to the PraFFL, validating the effectiveness of our proposed preference sampling adaptation method.

\noindent{\textbf{The Impact of Data Heterogeneity.}}
Table \ref{het} demonstrates that HetPFL consistently achieves the best performance in both local HV and global HV across all levels of data heterogeneity compared to seven baselines. Notably, the following observations emerge: (I) \textbf{Comprehensiveness}: First five baselines in Table \ref{het} fail to learn the entire Pareto front and struggle with high data heterogeneity. Their performance deteriorates as heterogeneity increases. HetPFL not only learns the entire Pareto front but also handles high heterogeneity effectively; (II) \textbf{Scalability}: When compared with personalized FL methods such as EquiFL and PraFFL, our proposed HetPFL excels in handling high data heterogeneity in both local and global datasets. EquiFL and PraFFL are better at handling high heterogeneity on local datasets, but they fail to address heterogeneity effectively on the global dataset.

\begin{table}[t]
    \centering
    \setlength{\abovecaptionskip}{0.06cm}
    \caption{Performance comparison across different heterogeneity levels on the SYNTHETIC dataset. The smaller the heterogeneity parameter, the greater the data heterogeneity.}\label{het}
    \setlength{\tabcolsep}{3pt}
    \begin{tabular}{lccccccc} 
        \toprule
        \multirow{3}{*}{Method} & \multicolumn{3}{c}{Local HV} & \multicolumn{3}{c}{Global HV} \\
        \cmidrule(lr){2-4} \cmidrule(lr){5-7} 
        & \multicolumn{3}{c}{{\small Heterogeneity Param.}} & \multicolumn{3}{c}{{\small Heterogeneity Param.}}\\
        & 0.1  & 5 & 1000 & 0.1  & 5 & 1000 \\ 
        \midrule
        LFT+Ensemble & 0.311  & 0.481 & 0.463 & 0.487 & 0.487  & 0.475 \\
        LFT+Fedavg & 0.593  & 0.692 & 0.701 & 0.468 & 0.468  & 0.468 \\
        Agnosticfair & 0.417  & 0.535 & 0.526 & 0.537 & 0.537  & 0.537 \\
        FairFed & 0.392  & 0.349 & 0.410 & 0.429 & 0.400  & 0.403 \\
        FedFB & 0.518  & 0.602 & 0.597 & 0.616 & 0.615  & 0.607 \\
        EquiFL & 0.734 & 0.626 & 0.615 & 0.584 & 0.644  & 0.643 \\
        PraFFL & \underline{0.802} & \underline{0.781} & \underline{0.789} & \underline{0.707} & \underline{0.744}  & \underline{0.768} \\
        HetPFL & \textbf{0.808} & \textbf{0.806} & \textbf{0.810} & \textbf{0.820} & \textbf{0.791} & \textbf{0.817} \\
        \bottomrule
    \end{tabular}
\end{table}

\begin{table}[t]
    \centering
    \setlength{\abovecaptionskip}{0.12cm}
    \setlength{\tabcolsep}{3pt}
    \caption{Performance comparison of different methods across the number of clients on SYNTHETIC dataset.}\label{clients}
    \begin{tabular}{lccccccc}
        \toprule
        \multirow{3}{*}{Method} & \multicolumn{3}{c}{Local HV} & \multicolumn{3}{c}{Global HV} \\
        \cmidrule(lr){2-4} \cmidrule(lr){5-7}
                & \multicolumn{3}{c}{Number of Clients} & \multicolumn{3}{c}{Number of Clients}\\
        & \multicolumn{1}{c}{10} & \multicolumn{1}{c}{100} & \multicolumn{1}{c}{300} & \multicolumn{1}{c}{10} & \multicolumn{1}{c}{100} & \multicolumn{1}{c}{300} \\
        \midrule
        LFT+Ensemble & 0.475  & 0.464 & 0.535 & 0.463 & 0.536  & 0.470 \\
        LFT+Fedavg & 0.664  & 0.472 & 0.605 & 0.460 & 0.589  & 0.469 \\
        Agnosticfair & 0.548  & 0.557 & 0.547 & 0.551 & 0.635  & 0.553 \\
        FairFed & 0.414  & 0.326 & 0.413 & 0.372 & 0.381  & 0.353 \\
        FedFB & 0.610  & 0.599 & 0.562 & 0.613 & \underline{0.667}  & 0.581 \\
        EquiFL & 0.564  & 0.602 & 0.620 & 0.613 & 0.573  & 0.607 \\
        PraFFL & \underline{0.803}  & \underline{0.716} & \underline{0.789} & \underline{0.807} & 0.578  & \underline{0.621} \\
        HetPFL & \textbf{0.808}  & \textbf{0.785} & \textbf{0.848} & \textbf{0.808} & \textbf{0.814}  & \textbf{0.813} \\
        \bottomrule
    \end{tabular}
\end{table}
\begin{table}[t]
    \centering
    \setlength{\tabcolsep}{6pt}
    \setlength{\abovecaptionskip}{0.12cm}
    \caption{Ablation experiments on preference sampling adaptation (PSA) and preference-aware hypernet fusion (PHF).}\label{ab}
    \begin{tabular}{c|cc|cc}
        \hline
       \rule{0pt}{10pt} Dataset & PSA & PHF & Local HV & Global HV \\
        \hline
        \multirow{4}{*}{SYNTHETIC}
        & \texttimes & \texttimes & 0.800 & 0.719 \\
        & \texttimes & \checkmark & 0.800 & 0.793 \\
        & \checkmark & \texttimes & 0.830 & 0.825 \\
        & \checkmark & \checkmark & \textbf{0.830} & \textbf{0.827} \\
         \bottomrule
        \multirow{4}{*}{COMPAS} 
        & \texttimes & \texttimes & 0.599 & 0.613 \\
        & \texttimes & \checkmark & 0.599 & \textbf{0.639} \\
        & \checkmark & \texttimes & 0.623 & 0.624 \\
        & \checkmark & \checkmark & \textbf{0.623} & {0.626} \\
         \bottomrule
        \multirow{4}{*}{BANK}
         & \texttimes & \texttimes & 0.901 & 0.895 \\
        & \texttimes & \checkmark & 0.901 & 0.896 \\
        & \checkmark & \texttimes & 0.904 & 0.886 \\
        & \checkmark & \checkmark & \textbf{0.904} & \textbf{0.898} \\
         \bottomrule
        \multirow{4}{*}{ADULT} 
        & \texttimes & \texttimes & 0.766 & 0.750 \\
        & \texttimes & \checkmark & 0.766 & 0.846 \\
        & \checkmark & \texttimes & 0.783 & 0.813 \\
        & \checkmark & \checkmark & \textbf{0.783} & \textbf{0.846} \\
        \hline
    \end{tabular}
\end{table}

\noindent{\textbf{The Impact of the Number of Clients.}}
We analyze Table \ref{clients} from following two aspects: (I) \textbf{Comprehensiveness}: The first six methods in Table \ref{clients}, which lack the capability to learn the Pareto front, exhibit limited performance in both small and large-scale client scenarios, with both local HV and global HV consistently below 0.7. In contrast, HetPFL not only learns the Pareto front but also achieves the best performance across all scenarios, with both local HV and global HV exceeding 0.78; (II) \textbf{Scalability}: Compared to PraFFL, HetPFL demonstrates clear advantages in large-scale client scenarios. Notably, PraFFL tends to collapse in learning the global Pareto front under large-scale settings, whereas HetPFL consistently achieves the best results in both local HV and global HV, regarding different client scales.

\noindent{\textbf{Ablation Study.}}
Table \ref{ab} reveals two key observations:
(I) PSA enhances the learning ability of the local Pareto front. On the SYNTHETIC dataset, the local HV improves from 0.80 to 0.83 with PSA. Similar improvements in local HV can be observed across the other three datasets.
(II) PHF enhances the performance of the global Pareto front. Without PSA, PHF achieves an approximately 8\% improvement in global HV on the SYNTHETIC dataset (from 0.719 to 0.793). With PSA, the improvement is smaller (i.e., 0.2\%) due to the high global HV had been achieved by PSA (i.e., 0.825). Similar patterns can be observed on the other three datasets.

\section{Conclusion}
In this paper, we proposed HetPFL, a comprehensive method for learning both local and global Pareto fronts in fair federated learning. First, HetPFL includes a Preference Sampling Adaptation (PSA) approach, which adaptively learns the preference sampling distribution for each client. Second, HetPFL incorporates a Preference-aware Hypernet Fusion (PHF) approach, which guides the generation of the global hypernet by learning the mapping from preferences to fusion weights at the server. We prove that HetPFL achieves an error convergence rate of order $\mathcal{O}(\frac{1}{t})$. Experimental results show that HetPFL outperforms seven state-of-the-art methods in learning both local and global Pareto fronts across four datasets.

\section*{Acknowledgments}
This work was supported in part by the National Natural Science Foundation of China under Grant 62202214 and Guangdong Basic and Applied Basic Research Foundation under Grant 2023A1515012819. 
\bibliographystyle{named}
\bibliography{ijcai25}
\clearpage


\appendix
\section{Schematic diagram}
\begin{figure}[h]
    \centering
    \includegraphics[width=\linewidth]{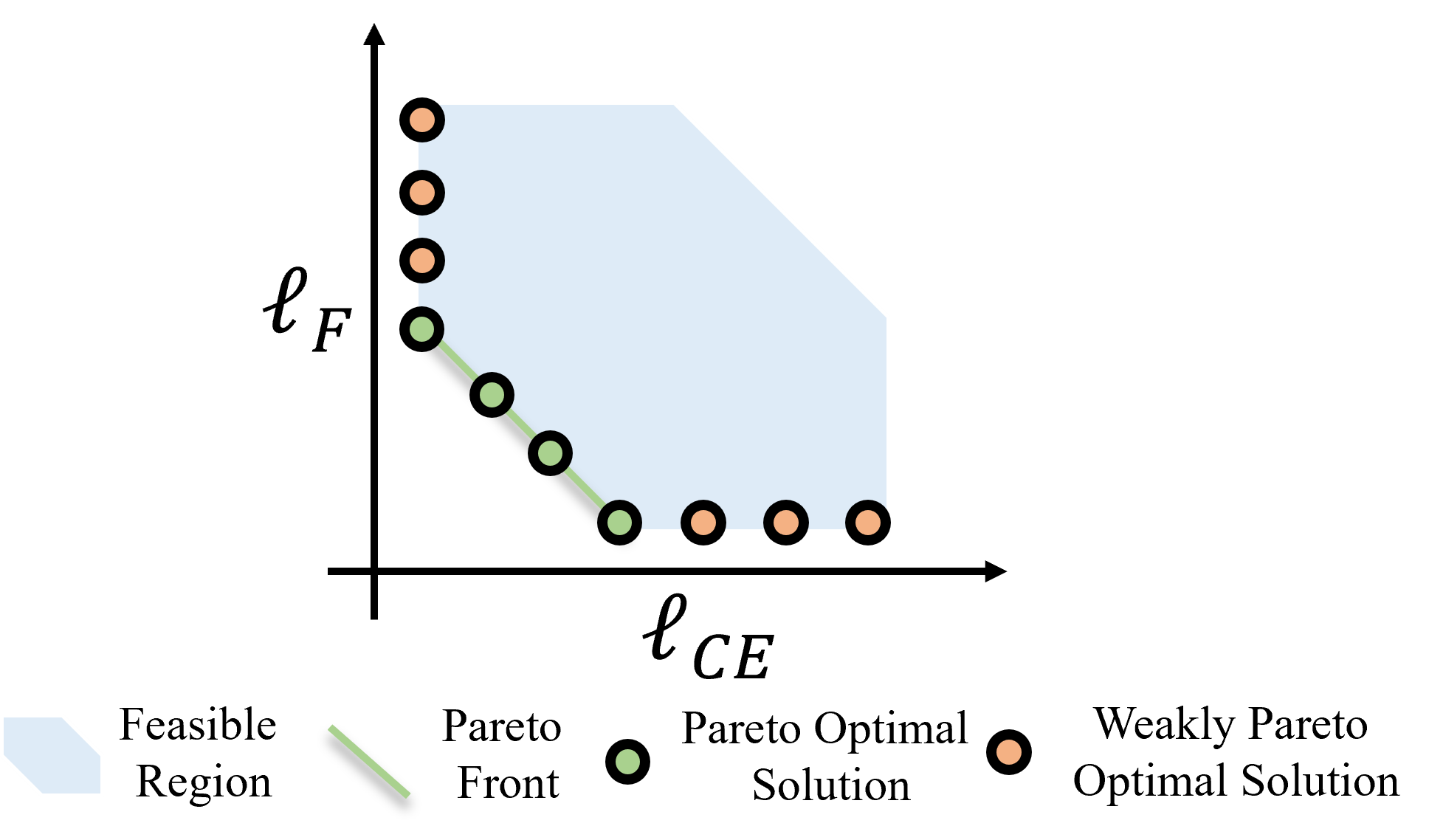}
    \caption{Schematic diagram of Pareto optimal solution, weakly Pareto optimal solution, and Pareto front.}
    \label{fig:def}
\end{figure}

Fig. \ref{fig:def} shows the concept of Pareto front, Pareto optimal solution and weakly Pareto optimal solution in the problem of minimizing two objectives. The green dots correspond to the Pareto optimal solutions, i.e., the optimal solution, under one trade-off. The green line corresponds to the Pareto front, which is the optimal solution under all trade-offs. The orange dots are weakly Pareto optimal solutions, which achieve the optimality in one objective while cannot guarantee the quality of the other objective.

\section{Proof of Theorem \ref{them}}
The proof logic proceeds as follows: Firstly, we give the upper bounds of the hypernet and the optimized sampling distribution within each iteration of a round (Lemma \ref{lemma2}). Next, we prove the convergence relationship between the hypernet and the communicated model at round $t$. Finally, since the hypernet and the optimization of sampling distribution are coupled, we further consider the influence of the sampling distribution to derive the upper bound of the hypernet at any given round $t$.

Before proving Theorem \ref{them}, we give the upper bounds of the hypernet and the optimized sampling distribution within each iteration of a round as follows:
\begin{lemma}[Convergence of the Hypernet and Sampling Distribution in One Round \cite{hong2023two}]\label{lemma2}
    If hypernet’s learning rate $\eta_h$ and sampling distribution’s learning rate $\kappa_s$ for any local iterations $s>0$ satisfy
    \begin{small} 
    \begin{equation}
        \kappa_s \! \leq\! \min\{c_1\eta^{\frac{3}{2}},\frac{1}{\mu_2}\}, \frac{\eta_{s-1}}{\eta_s}\! \leq \! 1+\frac{\eta_s\mu_1}{8}, \frac{\kappa_{s-1}}{\kappa_s}\! \leq \! 1+\frac{3\kappa_s \mu_2}{4},\nonumber
    \end{equation} 
    \end{small}
    \begin{small} 
    \begin{equation}
        \eta_s \leq \min \{ c_2\kappa_s^{\frac{2}{3}},\frac{1}{\mu_1},\frac{\mu_1}{L^2_{t1}\sigma_1^2}, \frac{\mu_1^2}{48c_1^2L^2L^2_q}\}, 8\mu_2\kappa_s \! \leq \! \mu_1\eta_s, \forall s \geq 0,\nonumber
    \end{equation}
    \end{small}
    where $c_1>0$, $c_2>0$,$ L=L_{h1}+\frac{L_{t4}G_1}{\mu_1}+G_2(\frac{L_{t2}}{\mu_1}+\frac{L_{h3}G_1}{\mu_1^2})$, $L_q$ are constants, then we have 
\begin{small}
\begin{equation}
\begin{aligned}\label{hyp}
    \Delta_{\boldsymbol{\alpha}_k}^s \leq &\left[\prod \limits_{j=0}^{s-1}(1-\kappa_j \mu_2)\right] 
    \left[\Delta_{\boldsymbol{\alpha}_k}^0+\frac{L^2}{\mu_2^2}\Delta_{\boldsymbol{\beta}_k}^0 \right]  \\ 
    &+ \frac{c_2L^2}{\mu_2^2} \left[\frac{\sigma_1^2}{\mu_1}+\frac{c_1^2 L_{q1}^2}{\mu_1}\sigma_3^2\right]{({\kappa}_{s-1})}^{\frac{2}{3}}+\frac{b_q^2}{\mu_2^2},
\end{aligned}
\end{equation}
\end{small}
\begin{small}
    \begin{equation}\label{alp}
        \Delta_{\boldsymbol{\beta}_k}^s \leq \left[\prod \limits_{j=0}^{s-1}(1-\frac{\eta_j\mu_1}{4})\right] \Delta_{\boldsymbol{\beta}_k}^0 + \left[\frac{\sigma_1^2}{\mu_1}+\frac{c_1^2L_{q1}^2}{\mu^2_1}\right]\eta_{s-1},
    \end{equation}
\end{small}
\!\!where $\sigma_1, \sigma_2$, and $\sigma_3$ are constants.
\end{lemma}

\begin{algorithm}[t] 
\caption{HetPFL algorithm}\label{alg}
\begin{algorithmic}[1] 
\STATE \textbf{Input:} $K, \mathcal{D}_{{k} \in [K]}, T, \tau_c, \tau_p,\eta_t, \eta_u, \eta_h, \kappa_h$, $N$, $U$, $\widetilde{\boldsymbol{\lambda}}$
\STATE \textbf{Output:} $\{\psi_{k}\}_{k \in [K]}$, $\{\boldsymbol{\beta}_{k}\}_{k \in [K]}$, $\boldsymbol{\psi}$, and $\boldsymbol{\varphi}$
\STATE \textit{// Each client initializes a hypernetwork $\boldsymbol{\beta}_{k}$}
\FOR{$t=1$ \TO $T$}
    \STATE \textit{// Server sends aggregated model $\psi_{k}$ to clients}
    \FOR{client $k \in [K]$ in parallel}
        \STATE \textit{// Optimize the communicated model}
        \FOR{$s=1$ \TO $\tau_{c}$}
            \STATE Update $\psi_{k}$ by Eq. (\ref{opcomm});
        \ENDFOR
        \STATE \textit{// Optimize the hypernet and sampling distribution}
        \FOR{$h=1$ \TO $\tau_{p}$}
            \STATE Sample the preference vector $\boldsymbol{\lambda} \in p(\boldsymbol{\alpha}_k)$;
            \STATE Update $\boldsymbol{\beta}_{k}$ by Eq. (\ref{eq10});
            \STATE Update $\boldsymbol{\alpha}_k$ by Eq. (\ref{opalpha});
        \ENDFOR
    \ENDFOR
    \STATE \textit{// Server aggregates the communicated model}
    \STATE $\boldsymbol{\psi}=\frac{1}{K}\sum_{k=1}^K \boldsymbol{\psi}_{k}$;
\ENDFOR
\STATE \textit{// Optimize the FusionNet $\boldsymbol{\varphi}$}
\FOR{$u=1$ \TO $U$}
\STATE Update $\boldsymbol{\varphi}$ by Eq. (\ref{opfusion});
\ENDFOR
\STATE \textbf{Return:} $\{\boldsymbol{\psi}_{k}\}_{k \in [K]}$, $\{\boldsymbol{\beta}_{k}\}_{k \in [K]}$, $\boldsymbol{\psi}$, and $\boldsymbol{\varphi}$
\end{algorithmic}
\end{algorithm}
We further analyze the convergence relationship between the hypernet and the communicated model across different rounds. Among them, we incorporate the impact of Eqs. (\ref{hyp}) and (\ref{alp}) on the convergence of hypernet $\mathbb{E}\left[\|\boldsymbol{\beta}_k^{t+1}-\boldsymbol{\beta}_k^*\|^2\right]$ at round $t+1$.
\begin{align}
\mathbb{E}&\left[\|\boldsymbol{\beta}_k^{t+1}-\boldsymbol{\beta}_k^*\|^2\right]& \nonumber \\= & \ \mathbb{E}\left[\|\boldsymbol{\beta}_k^t-\eta_t \nabla_{\boldsymbol{\beta}} {g}(\boldsymbol{\beta}_k^t,\boldsymbol{\psi}^t)-\boldsymbol{\beta}_k^*\|^2\right] \\
=& \mathbb{E}\left[\|\boldsymbol{\beta}_k^t-\boldsymbol{\beta}_k^*\|^2\right] + \eta_t^2 \mathbb{E}\left[\|I^{t}_{k}\nabla_{\boldsymbol{\beta}} {g}(\boldsymbol{\beta}_k^t,\boldsymbol{\psi}^t)\|^2\right] \nonumber\\
&+ 2\eta_t \mathbb{E}\left[p_{k}{\nabla_{\boldsymbol{\beta}} {g}(\boldsymbol{\beta}_k^t,\boldsymbol{\psi}^t)}^{T}(\boldsymbol{\beta}_k^*-\boldsymbol{\beta}_k^t)\right] \\
\overset{(a)}{\leq}&(1-\eta_t {L}_{t1})\mathbb{E}\left[\|\boldsymbol{\beta}_k^t-\boldsymbol{\beta}_k^*\|^2\right] + \underbrace{\eta_t^2\mathbb{E}\left[\|\nabla_{\boldsymbol{\beta}} {g}(\boldsymbol{\beta}_k^t;\boldsymbol{\psi}^t)\|^2\right]}_{\mathcal{C}_1} \nonumber\\
&+\underbrace{2\eta_t  \mathbb{E}\left[{g}(\boldsymbol{\beta}_k^*,\boldsymbol{\psi}^t)-{g}_(\boldsymbol{\beta}_k^t,\boldsymbol{\psi}^t)\right]}_{\mathcal{C}_2},\label{total}
\end{align}
where $(a)$ is due to the smoothness of $g$. We then find the upper bounds of $\mathcal{C}_1$ and $\mathcal{C}_2$, respectively. For $\mathcal{C}_1$,

\begin{align} 
\eta_t^2 &\mathbb{E}\left[\|\nabla_{\boldsymbol{\beta}} {g}(\boldsymbol{\beta}_{k}^{t},\boldsymbol{\psi}^t)\|^2\right]  \nonumber\\ =& \eta_t^2 \mathbb{E}\left[\|\nabla_{\boldsymbol{\beta}} {g}(\boldsymbol{\beta}_{k}^{t},\boldsymbol{\psi}^t)\!-\!\nabla_{\boldsymbol{\beta}} {g}(\boldsymbol{\beta}_{k}^{t},\boldsymbol{\psi}^*)\! +\! \nabla_{\boldsymbol{\beta}} {g}(\boldsymbol{\beta}_{k}^{t},\boldsymbol{\psi}^*)\|^2\right]
\\
=& 2\eta_t^2 \mathbb{E}\left[\|\nabla_{\boldsymbol{\beta}} {g}(\boldsymbol{\beta}_{k}^{t},\boldsymbol{\psi}^*)\| \|\nabla_{\boldsymbol{\beta}} {g}(\boldsymbol{\beta}_{k}^{t},\boldsymbol{\psi}^t)-\nabla_{\boldsymbol{\beta}} {g}(\boldsymbol{\beta}_{k}^{t},\boldsymbol{\psi}^*)\|\right]  \nonumber\\
&\!+\!\eta_t^2 \mathbb{E}\left[\|\nabla_{\boldsymbol{\beta}} {g}(\boldsymbol{\beta}_{k}^{t},\boldsymbol{\psi}^*)\|^2\right]\nonumber \\ &+\eta_t^2 \mathbb{E}\left[\|\nabla_{\boldsymbol{\beta}} {g}(\boldsymbol{\beta}_{k}^{t},\boldsymbol{\psi}^t)-\nabla_{\boldsymbol{\beta}} {g}(\boldsymbol{\beta}_{k}^{t},\boldsymbol{\psi}^*)\|^2\right]\\
\overset{(b)}{\leq} &\eta_t^{2} \mathbb{E}\left[\|\nabla_{\boldsymbol{\beta}} {g}(\boldsymbol{\beta}_{k}^{t}, \boldsymbol{\psi}^*)\|^{2}\right]\nonumber \\ &+ 2 \eta_t^{2} L_{t1} \mathbb{E}\left[\|\nabla_{\boldsymbol{\beta}} {g}(\boldsymbol{\beta}_{k}^{t}, \boldsymbol{\psi}^*)\| \|\boldsymbol{\psi}^t - \boldsymbol{\psi}^*\|\right]  \nonumber\\
&+\eta_t^{2}\check{L}^2\mathbb{E}\left[\|\boldsymbol{\psi}^t-\boldsymbol{\psi}^*\|^{2}\right],\label{up1}
\end{align}
where $(b)$ is due to the convexity of $g$. Then, we continue to find the upper bound of $\mathcal{C}_{2}$,
\begin{align}
    2 \eta_t & \mathbb{E}\left[{g}(\boldsymbol{\beta}_k^*, \boldsymbol{\psi}^t) - {g}(\boldsymbol{\beta}_k^t, \boldsymbol{\psi}^t)\right] \nonumber \\
    \overset{(c)}{\leq} 2 &\eta_t  \mathbb{E}\left[{g}(\boldsymbol{\beta}_k^*, \boldsymbol{\psi}^t) - {g}(\boldsymbol{\beta}_k^t, \boldsymbol{\psi}^t) \right] \nonumber \\
     + &\eta_t (L_{t1} - \mu_1) \|\boldsymbol{\psi}^t - \boldsymbol{\psi}^*\|^2 \nonumber \\
    + & \mathbb{E}\left[\left(\nabla_{\boldsymbol{\beta}} {g}(\boldsymbol{\beta}_k^*, \boldsymbol{\psi}^t) - \nabla_{\boldsymbol{\beta}} {g}(\boldsymbol{\beta}_k^t, \boldsymbol{\psi}^t)\right)^{\mathsf{T}} (\boldsymbol{\psi}^t - \boldsymbol{\psi}^*)\right] \nonumber \\
    \overset{(d)}{\leq} 2 &\eta_t  \mathbb{E}\left[{g}(\boldsymbol{\beta}_k^*, \boldsymbol{\psi}^t) - {g}(\boldsymbol{\beta}_k^t, \boldsymbol{\psi}^t) \right] \nonumber \\
     + &\eta_t  (L_{t1} - \mu_1) \|\boldsymbol{\psi}^t - \boldsymbol{\psi}^*\|^2 \nonumber \\
    + & 2\eta_t  L_{t1} \mathbb{E}\left[\|\boldsymbol{\phi}_k^t - \boldsymbol{\phi}_k^*\| \|\boldsymbol{\psi}_k^t - \boldsymbol{\psi}_k^*\|\right], \label{up2}
\end{align}
where $(c)$ uses the smoothness and convexity of $g$. $(d)$ uses the convexity of $g$. So far, we have found the upper bounds of $\mathcal{C}_{1}$ and $\mathcal{C}_{2}$. We then plugging their upper bounds Eq. (\ref{up1}) and Eq. (\ref{up2}) into Eq. (\ref{total}), as follows:
\begin{align}
\mathbb{E}&\left[\|\boldsymbol{\beta}_k^{t+1}-\boldsymbol{\beta}_k^*\|^2\right]\nonumber\\ \leq
&(1-\eta_t \mu_1) \mathbb{E}\left[\|\boldsymbol{\beta}_k^t-\boldsymbol{\beta}_k^*\|^2\right] + \eta_t^{2} \mathbb{E}\left[\|\nabla_{\boldsymbol{\beta}} {g}_(\boldsymbol{\beta}_{k}^{t}, \boldsymbol{\psi}^*)\|^{2}\right] \nonumber\\
&+2\eta_t^{2} L_{t1} \mathbb{E}\left[\|\nabla_{\boldsymbol{\beta}} {g}(\boldsymbol{\beta}_{k}^{t}, \boldsymbol{\psi}^*)\| \|\boldsymbol{\psi}^t-\boldsymbol{\psi}^*\|\right] \nonumber\\
&+ \eta_t^{2} L_{t1}^2 \mathbb{E}\left[\|\boldsymbol{\psi}^t-\boldsymbol{\psi}^*\|^{2}\right]+2\eta_t \mathbb{E}\left[{g}(\boldsymbol{\beta}_k^*,\boldsymbol{\psi}^t) - {g}(\boldsymbol{\beta}_k^t,\boldsymbol{\psi}^t) \right] \nonumber\\
&+2\eta_t L_{t1} \mathbb{E}\left[ \|\boldsymbol{\beta}_k^t - \boldsymbol{\beta}_k^*\| \|\boldsymbol{\psi}^t - \boldsymbol{\psi}^*\| \right]\nonumber\\
&+ \eta_t (L_{t1} - \mu_1) \mathbb{E}\left[\|\boldsymbol{\psi}^t - \boldsymbol{\psi}^*\|^2\right]  \\ \overset{(e)}{\leq} &(1-\eta_t  \mu_1) \mathbb{E}\left[\|\boldsymbol{\beta}_k^t - \boldsymbol{\beta}_k^*\|^2\right] + \eta_t^{2} \mathbb{E}\left[\|\nabla_{\boldsymbol{\beta}} {g}(\boldsymbol{\beta}_{k}^{t}, \boldsymbol{\psi}^*)\|^{2}\right]\nonumber\\
&+2\eta_t^{2} L_{t1} \sqrt{\mathbb{E}\left[\|\nabla_{\boldsymbol{\beta}} {g}(\boldsymbol{\beta}_{k}^{t}, \boldsymbol{\psi}^*)\|^2\right] \mathbb{E}\left[\|\boldsymbol{\psi}^t-\boldsymbol{\psi}^*\|^2\right]} \nonumber\\
&+\eta_t^{2} L_{t1}^2 \mathbb{E}\left[\|\boldsymbol{\psi}^t-\boldsymbol{\psi}^*\|^2\right] +\eta_t  (L_{t1} - \mu_1) \mathbb{E}\left[\|\boldsymbol{\psi}^t - \boldsymbol{\psi}^*\|^2\right]\nonumber\\
&+2\eta_t L_{t1} \sqrt{\mathbb{E}\left[\|\boldsymbol{\beta}_k^t - \boldsymbol{\beta}_k^*\|^2\right] \mathbb{E} \left[\|\boldsymbol{\psi}^t - \boldsymbol{\psi}^*\|^2\right]}  \\
\overset{(f)}{\leq} &(1-\eta_t \mu_1) \mathbb{E}\left[\|\boldsymbol{\beta}_k^t - \boldsymbol{\beta}_k^*\|^2\right] +  2\eta_t^2 G_3 L_{t1} \sqrt{\mathbb{E}\left[\|\boldsymbol{\psi}^t - \boldsymbol{\psi}^*\|^2\right]} \nonumber\\ &+\eta_t^{2} G_3^2 + (\eta_t^{2} L_{t1}^2 + \eta_t  (L_{t1} - \mu_1)) \mathbb{E}\left[\|\boldsymbol{\psi}^t - \boldsymbol{\psi}^*\|^2\right] \nonumber\\
&+ 2\eta_t  L_{t1} \sqrt{\mathbb{E}\left[\|\boldsymbol{\beta}_k^t - \boldsymbol{\beta}_k^*\|^2\right] \mathbb{E} \left[\|\boldsymbol{\psi}^t - \boldsymbol{\psi}^*\|^2\right]}, \label{sa}
\end{align}
where $(e)$ is due to the Cauchy-Schwarz inequality and $\mathbb{E}\big[{g}(\boldsymbol{\beta}_k^*;\boldsymbol{\psi}^t)-{g}(\boldsymbol{\beta}_k^t;\boldsymbol{\psi}^t)\big]\leq 0$. $(f)$ holds because the norm of the squared gradient is bounded by $G_3^2$.

Eq. (\ref{sa}) describes the convergence relationship between the hypernet and the communicated model at each round. To further consider the impact of sampling distribution, we substitute Lemma \ref{commconv} and Lemma \ref{lemma2} into the right side of the inequality Eq. (\ref{sa}). Then, we denote the convergence error of hypernet at round $t+1$ as $\Delta_{\boldsymbol{\beta}_k}^{t+1}=\mathbb{E}\left[\|\boldsymbol{\beta}_k^{t+1}-\boldsymbol{\beta}_k^*\|^2\right]$, its has the following upper bound:
        \begin{small}
        \begin{equation}
        \begin{aligned}
            \Delta_{\boldsymbol{\beta}_k}^{t+1} \leq (1&-\mu_1\eta_t) \left[(\frac{3}{4})^{\tau_p t} \Delta_{\boldsymbol{\beta}_k}^0 + \left[\frac{\sigma_1^2}{\mu_1^2}+\frac{c_1^2L_{q1}^2}{\mu^3_1}\right]\right]
            \\ +& 
            2\eta_t^2G_3L_{t1}(1-\eta_t\zeta)^{\frac{t}{4}}\sqrt{\Delta_{\boldsymbol{\psi}_k}^0} + \eta_t^2G_3^2
            \\ +&
            (\eta_t^2L_{t1}^2+\eta_t(L_{t1}-\mu_1))(1-\eta_t\zeta)^{\frac{t}{2}}\Delta^0_{\boldsymbol{\psi}_k}
            \\ +&
            2\eta_tL_{t1}(1-\eta_t\zeta)^{\frac{t}{4}}\sqrt{\Delta^0_{\boldsymbol{\psi}_k}}(\sqrt{\Delta_{\boldsymbol{\beta}_k}^0}+\sqrt{\frac{\sigma_1^2}{\mu_1^2}+\frac{c_1^2L_{q1}^2}{\mu^3_1}}).
            \nonumber
            \end{aligned}
        \end{equation}
    \end{small}
\noindent{Let} $z_1=2\eta_2L_{t1}+\sqrt{\frac{\sigma_1^2\mu_1+c_1^2L^2_{q1}}{\mu_1^3}}$ and $z_2=\eta_t^2L_{t1}^2+\eta_t(L_{t1}-\mu_1$. we get the following inequality:
        \begin{small}
        \begin{equation}
        \begin{aligned}
            \Delta_{\boldsymbol{\beta}_k}^{t+1} \leq 
            & \ (\frac{3}{4})^{\tau_p t}\Delta_{\boldsymbol{\beta}_k}^0 + z_1(1-\eta_t\zeta)^{t/4}\sqrt{\Delta^0_{\boldsymbol{\psi}_k}}
            \\ &+
            z_2(1-\eta_t\zeta)^{t/2}\Delta^0_{\boldsymbol{\psi}_k}+ 
    \frac{\sigma_1^2\mu_1+c_1^2L^2_{q1}+G_3^2\mu_1}{\mu^3_1}
            \\ &+
            2\eta_tL_{t1}(1-\eta_t\zeta)^{t/4}\sqrt{\Delta^0_{\boldsymbol{\psi}_k}\Delta^0_{\boldsymbol{\beta}_k}}.
            \end{aligned}
        \end{equation}
    \end{small}

\section{Additional Experiments}
\subsection{Experimental Details}
\subsubsection{Experimental Parameters}\label{param}
According to \cite{zeng2021improving}, we set the local epoch in each round to be 30 (i.e., $\tau_p+\tau_c=30$). The total communication rounds $T$ is set to 15, the batch size of data is 128 in the neural network training, and the batch size of sampling preference vectors $N$ is 4. We use Adam as optimizer for communicated model, hypernet, and preference distribution optimization. The reference point $\boldsymbol{r}$ in calculating hypervolume is set to (1, 1). The preference sampling distribution $p(\boldsymbol{\alpha}_k)$, $k \in [K]$, is represented by Dirichlet distribution. 
\subsubsection{Dataset}
According to the setting of \cite{zeng2021improving,ezzeldin2023fairfed}, the COMPAS and ADULT datasets have two clients, while the SYNTHETIC and BANK have three clients. This setting is common in cross-silo FL. In addition, we also conducted experiments with large-scale clients (e.g., 300 clients) in the experimental section.

The information of the four datasets is as follows:
\begin{itemize}
\item \textbf{SYNTHETIC} \cite{zeng2021improving} is an artificially generated dataset containing 5000 samples with two non-sensitive features and a binary sensitive feature;
\item  \textbf{COMPAS} \cite{barenstein2019propublica}: contains 7214 samples, 8 non-sensitive features,
and 2 sensitive features (i.e., gender and race);
\item \textbf{BANK} \cite{moro2014data} contains 45211 samples, 13 non-sensitive features,
and one sensitive feature (i.e., age);
\item \textbf{ADULT} \cite{dua2017uci} contains 48842 samples, with 13 non-sensitive features
and one sensitive feature (i.e., gender).
\end{itemize}
In the three real-world datasets, we split 30\% of the data into the test set, and the other 70\% of the data are split into the training set and validation set with a ratio of 9:1. Furthermore, we select the model that performs best in the validation set for testing.
\begin{figure}[t]
    \centering
    \includegraphics[width=0.82\linewidth]{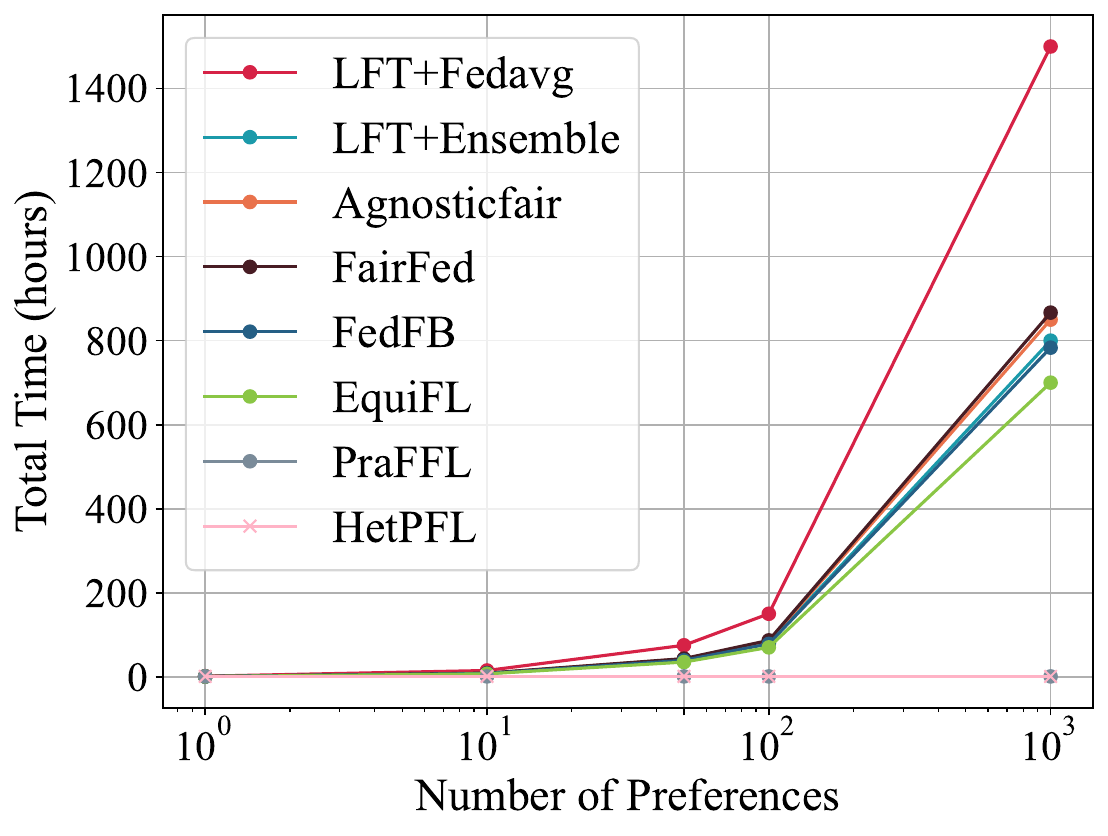}
    \caption{Comparison of the training time of the methods as the number of client preferences increases on the COMPAS dataset.}
    \label{time}
\end{figure}

\subsubsection{Baselines}
The introduction of seven baselines is as follows:
\begin{itemize}
    \item \textbf{LFT+Ensemble} learns fair classifiers from decentralized data without violating privacy policies by training locally fair classifiers and assembling them;
    \item \textbf{LFT+Fedavg} \cite{zeng2023federated} applies FedAvg \cite{mcmahan2017communication}, under which the server periodically aggregates and synchronizes locally trained models;
    \item \textbf{Agnosticfair} \cite{du2021fairness} uses kernel reweighing functions to assign a reweighing value
on each training sample in both loss function and fairness constraint;
    \item \textbf{FairFed} \cite{ezzeldin2023fairfed} enhances group fairness via a fairness-aware aggregation method;
    \item \textbf{FedFB} \cite{zeng2021improving} uses bi-level optimization to handle fairness. The lower layer is designed to optimize model performance and the upper layer is designed to improve model fairness;
    \item \textbf{EquiFL} \cite{makhija2024achieving} incorporates a
fairness term and personalized layer into the local optimization, effectively balancing local performance and fairness;
    \item \textbf{PraFFL} \cite{ye2024praffl} design a method that can provide a preference-specific model in real time based on the client’s preferences, which has the ability to learn the entire Pareto front.
\end{itemize}
\subsubsection{Neural Network Architecture.}
The architecture of the communicated model $\boldsymbol{\psi}$ is as follow:
\begin{equation}
    \begin{aligned}
\boldsymbol{\psi}(\boldsymbol{x})\!:\! \boldsymbol{x} & \!\to\! \text{Linear}(|\boldsymbol{x}|,4)\! \to\! \text{ReLU} \\
&\to\mathrm{Linear}(4,4)\to\mathrm{ReLU} \to \boldsymbol{x}_{\mathrm{mid}},
\end{aligned}
\end{equation}
where $\boldsymbol{x}$ represent features of dataset.

The architecture of hypernet
$h_{\boldsymbol{\beta}_k}$ is as follow:
\begin{equation}
    \begin{aligned}
h_{\boldsymbol{\beta}_k}:\boldsymbol{\lambda} & \to\text{Linear}(|\boldsymbol{\lambda}|,4) \to \text{ReLU} \\
&\to \mathrm{Linear}(4,4) \to \text{ReLU} \\
&\to \mathrm{Linear}(4,|\boldsymbol{\phi}_k|),
\end{aligned}
\end{equation}
where $\boldsymbol{\lambda}$ is preference vector, and the hypernetwork maps the preference vector to the personalized model $\boldsymbol{\phi}_k$.  

The architecture of FusionNet $W_{\boldsymbol{\varphi}}$ is as follows:
\begin{equation}
    \begin{aligned}
W_{\boldsymbol{\varphi}}:\boldsymbol{\lambda} & \to\text{Linear}(|\boldsymbol{\lambda}|,4) \to \text{ReLU} \\
&\to \mathrm{Linear}(4,4) \to \text{ReLU} \\
&\to \mathrm{Linear}(4,K),
\end{aligned}
\end{equation}
where $K$ is the number of clients.
\begin{figure}[t]
    \centering
    \includegraphics[width=0.49\linewidth,page=1]{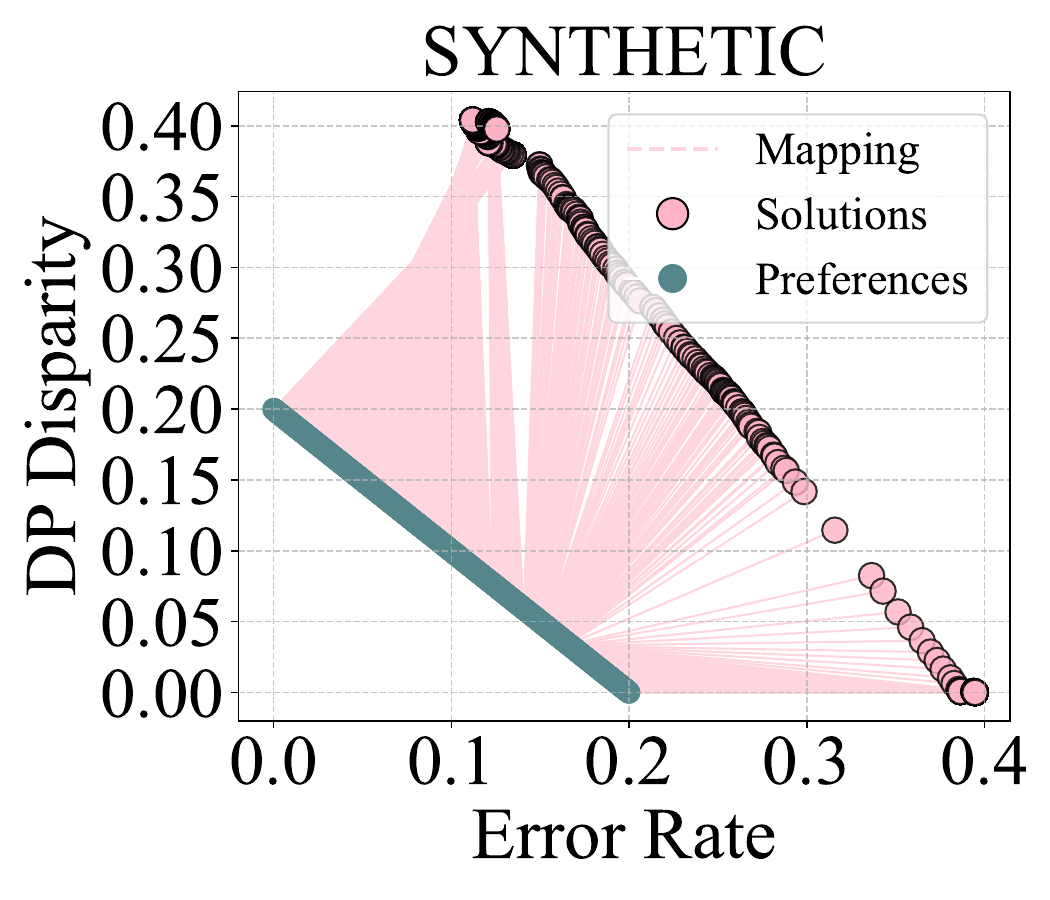} 
    \includegraphics[width=0.49\linewidth,page=1]{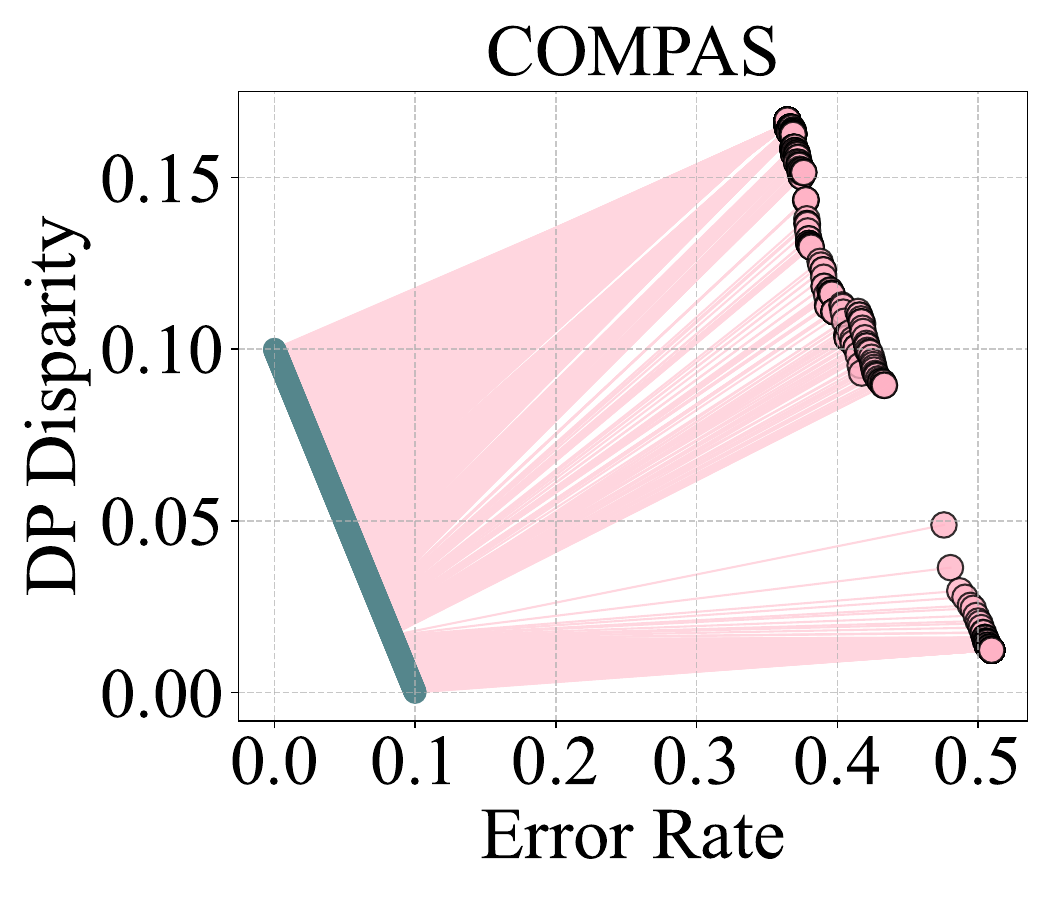} 
    \\
    \includegraphics[width=0.49\linewidth,page=1]{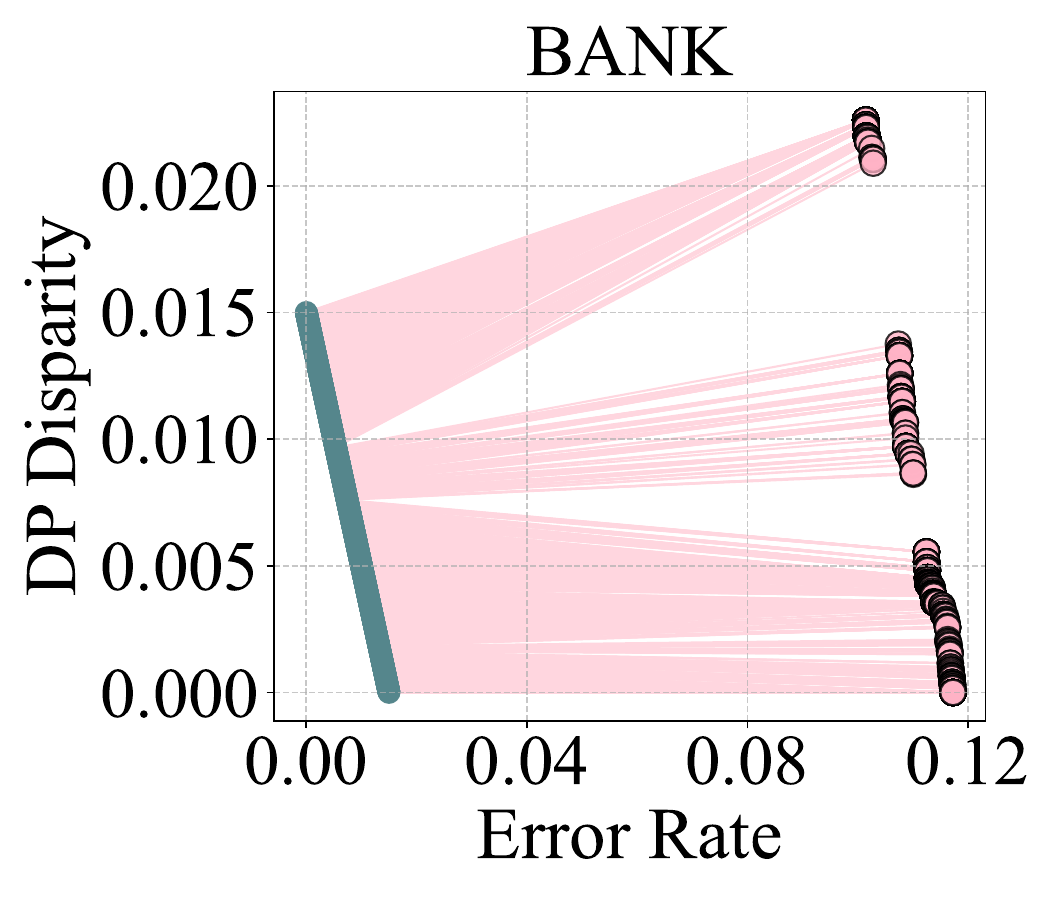} 
    \includegraphics[width=0.49\linewidth,page=1]{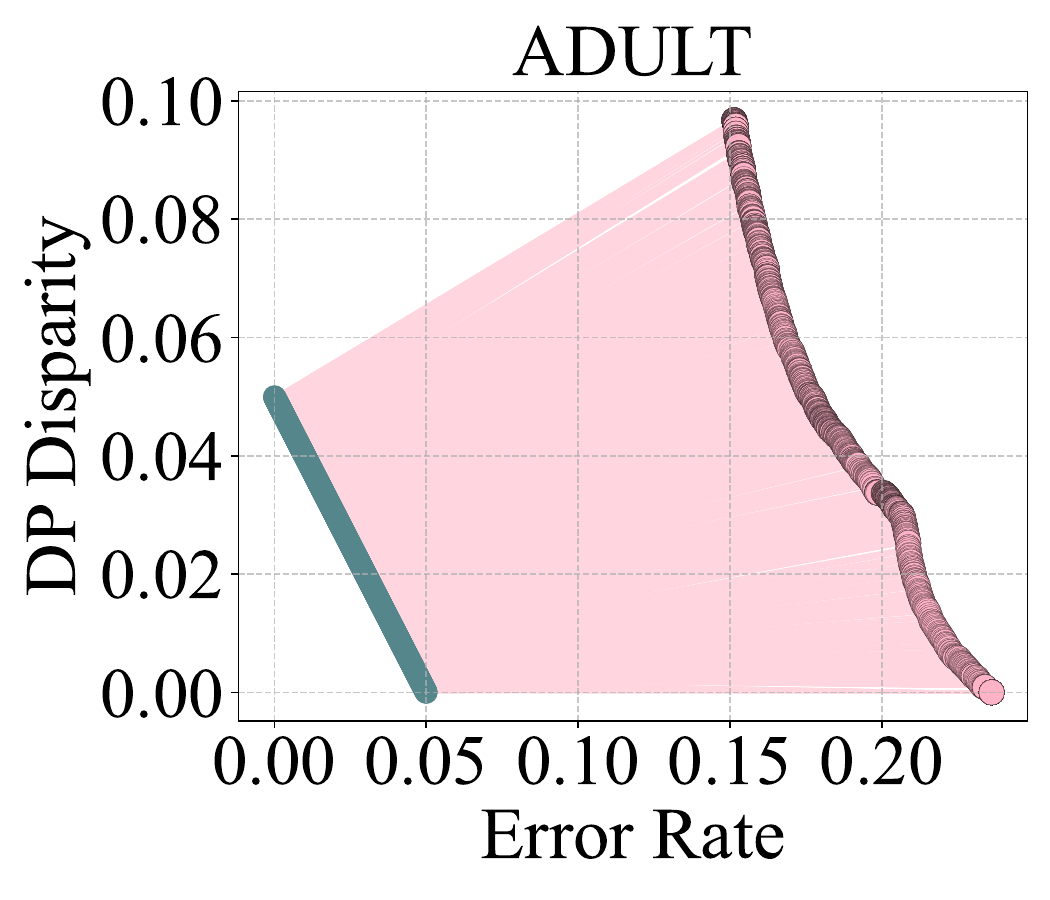} 
        \caption{Visualization of HetPFL’s mapping from the preferred sampling space to the objective space on the global dataset.}
    \label{conv}
\end{figure}
\begin{figure}[!t]
    \centering
    \begin{subfigure}{\linewidth}
        \centering
        \includegraphics[width=0.7\linewidth]{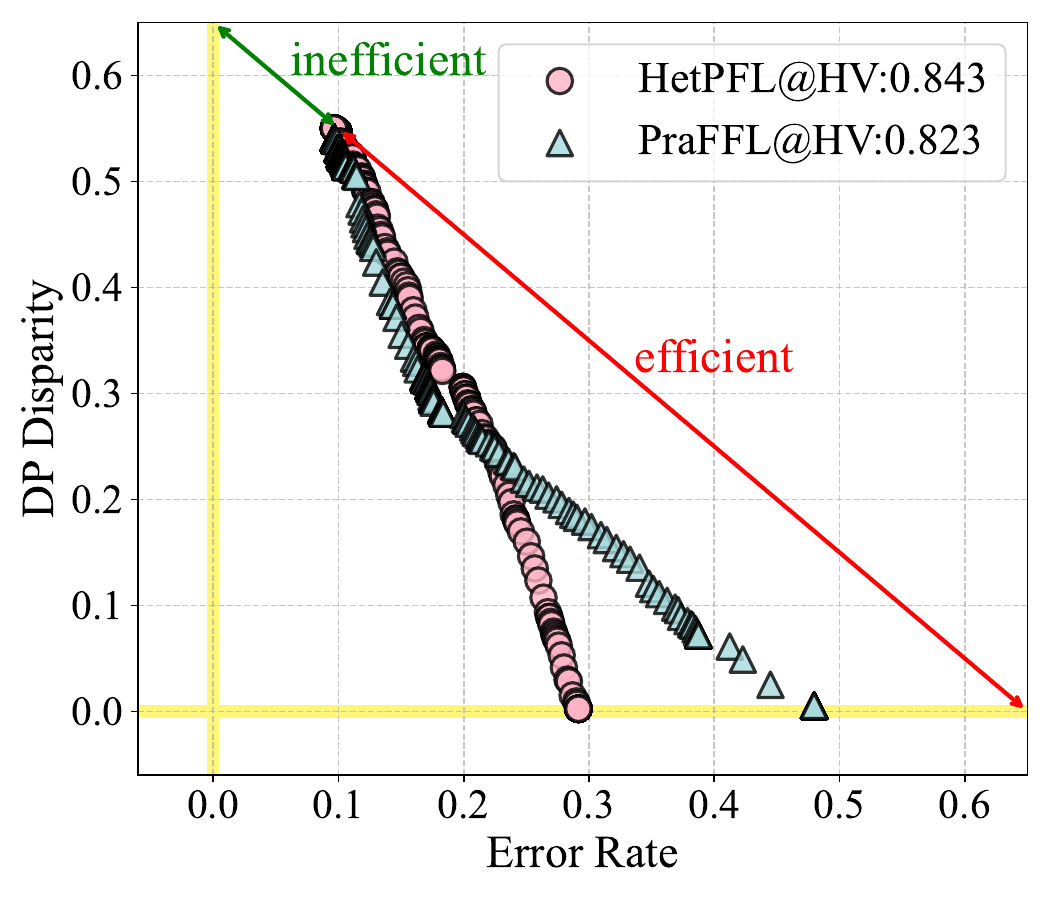}
        \caption{The local Pareto front learned by a client on the SYNTHETIC.}
    \end{subfigure}

    \begin{subfigure}{\linewidth}
        \centering
        \includegraphics[width=0.7\linewidth]{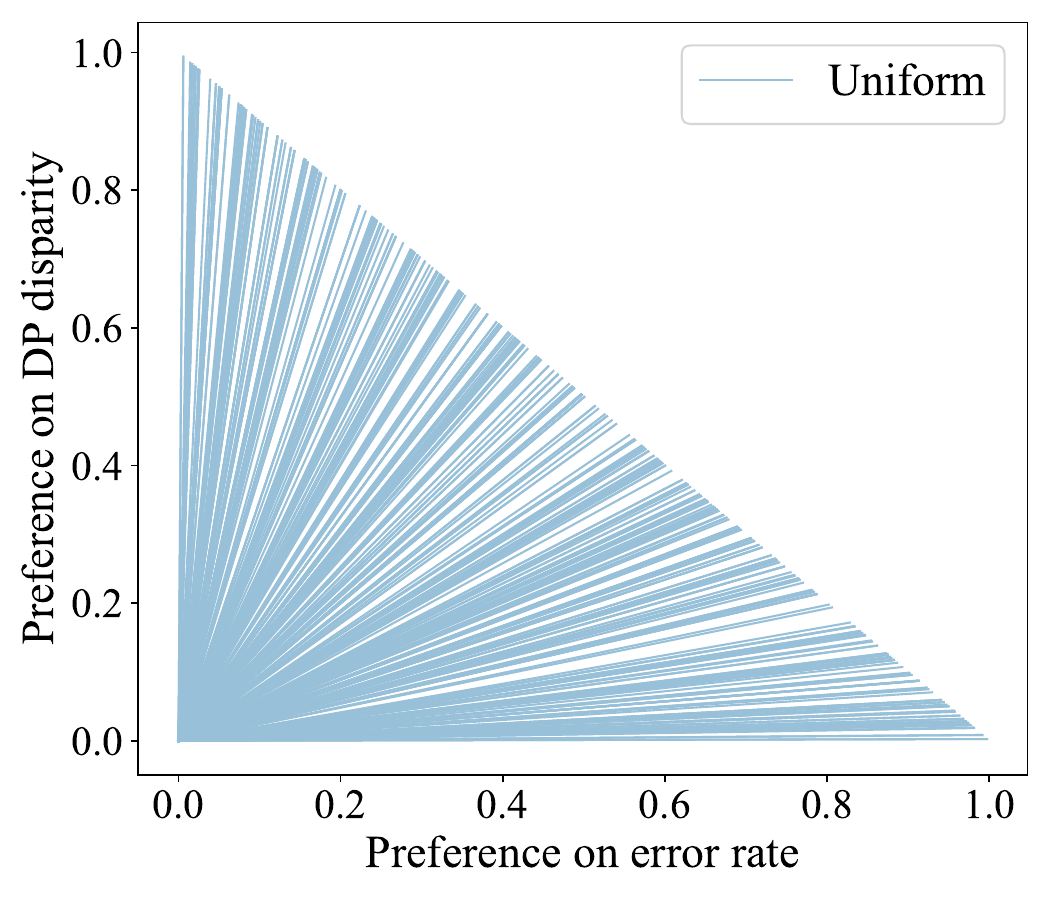}
        \caption{PraFFL's preference sampling distribution.}
    \end{subfigure}

    \begin{subfigure}{\linewidth}
        \centering
        \includegraphics[width=0.7\linewidth]{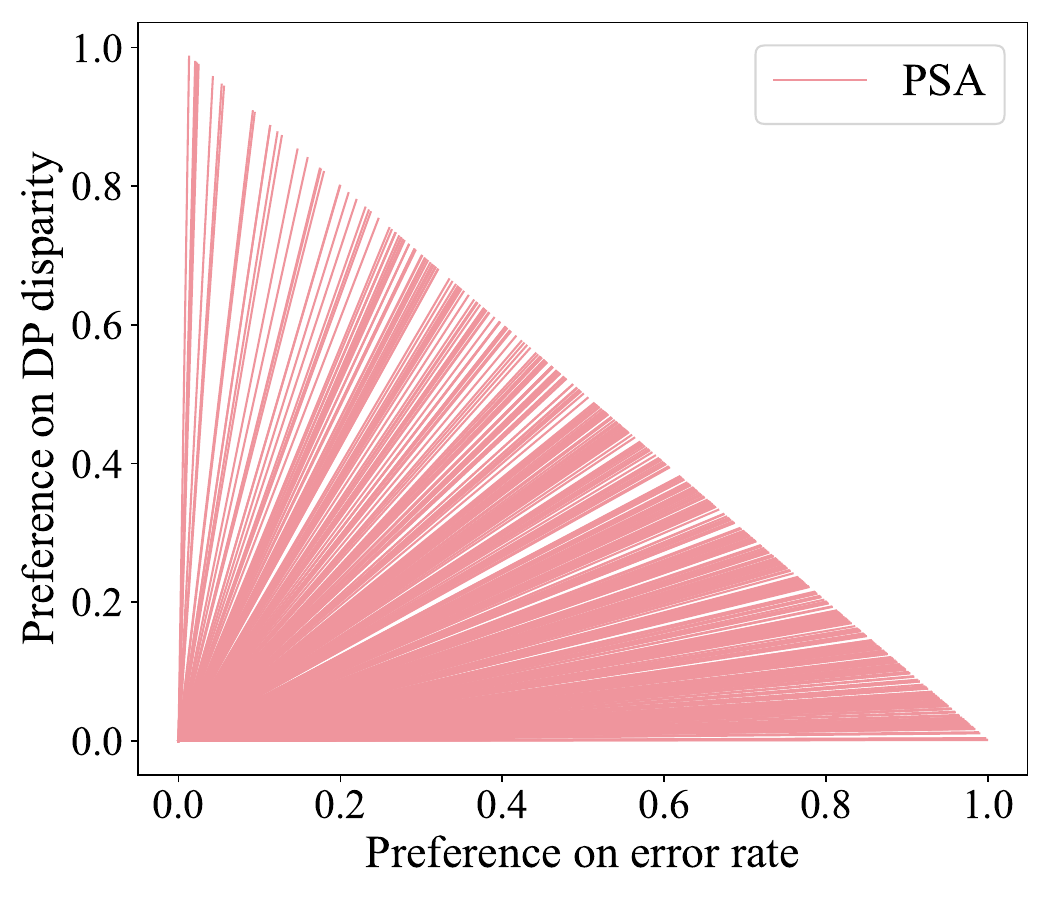}
        \caption{HetPFL's preference sampling distribution.}
    \end{subfigure}
    \caption{Comparison of Preference Sampling Distributions.}\label{dist}
\end{figure}
\subsection{Training Complexity under Different Preference Scales}
Fig. \ref{time} illustrates the training complexity of different methods for large-scale preferences. For methods capable of learning the Pareto front like PraFFL and HetPFL, they have learned the mapping from client preferences to corresponding models during the training phase. As a result, during the inference phase, they only need to substitute client’s preferences to generate corresponding preference-specific models. Therefore, their time complexity is $O(1)$. In contrast, for methods that cannot learn the Pareto front, their training complexity increases linearly with the number of client preferences.

\subsection{Visualization of Mapping}
Fig. \ref{conv} shows the mapping results of 1000 preference vectors by HetPFL across four datasets. To enhance visual clarity, we scale the preference sampling space. Each pink line represents a mapping from a preference to its corresponding model’s performance and fairness. It can be observed that HetPFL can map preference vectors to the Pareto front. Notably, on the ADULT dataset, HetPFL achieves particularly favorable results, as the mapping from preference vectors to the Pareto front is quite dense. On the SYNTHETIC dataset, some intersecting mapping paths appear mainly because during the optimization phase, the objective is formulated using loss functions, which differs from the error rate and DP disparity presented in the figure.

\subsection{Visualization of Preference Sampling Distribution}
Fig. \ref{dist} shows the Pareto fronts on a client and sampling distributions learned by PraFFL and HetPFL on the client using the SYNTHETIC dataset. We observe that the preference vector pointing to the green arrow area has no intersection with the Pareto front, while the preference vector pointing to the red arrow area has an intersection with the Pareto front. According to Proposition \ref{prop}, sampling preferences from the green area is inefficient for learning the Pareto front. Instead, an effective preference sampling distribution should focus more on the directions where the Pareto front is located (i.e., red area). Fig. \ref{dist} (c) shows the sampling distribution of HetPFL. We can clearly observe that the density is higher on the direction of the Pareto front. Therefore, the sampling distribution learned by HetPFL is more advantageous than the uniform sampling distribution used by PraFFL (see Fig. \ref{dist} (b)).
\end{document}